\documentclass[11pt]{article}

\usepackage[preprint]{acl}

\usepackage{times}
\usepackage{latexsym}
\usepackage[T1]{fontenc}
\usepackage[utf8]{inputenc}
\usepackage{microtype}
\usepackage{inconsolata}
\usepackage{graphicx}
\usepackage{amsmath}
\usepackage{amssymb}
\usepackage{amsthm}
\usepackage{booktabs}
\usepackage{placeins}
\usepackage{float}
\usepackage{tikz}
\usetikzlibrary{arrows.meta,positioning,fit,backgrounds}

\theoremstyle{definition}
\newtheorem{definition}{Definition}
\newtheorem{example}{Example}

\newcommand{\Ent}{\mathrm{Ent}}
\newcommand{\Con}{\mathrm{Con}}
\newcommand{\Neu}{\mathrm{Neu}}
\newcommand{\Inv}{\mathrm{Invalid}}
\newcommand{\Lworld}{L}
\newcommand{\Wlocal}{\mathcal{W}_{1}}
\DeclareMathOperator*{\argmax}{arg\,max}

\title{Learning Proposition-Level Links from Sentence-Level\\
Supervision for Neuro-Symbolic NLI}

\author{Xuyao Feng\textsuperscript{*} \and Antonis Bikakis \\
  University College London, United Kingdom \\
  \texttt{\{xuyao.feng.20,a.bikakis\}@ucl.ac.uk} \\
  \textsuperscript{*}Corresponding author.}

\hypersetup{
  pdftitle={Learning Proposition-Level Links from Sentence-Level Supervision for Neuro-Symbolic NLI},
  pdfauthor={Xuyao Feng and Antonis Bikakis}
}

\begin{document}
\maketitle

\begin{abstract}
We study neuro-symbolic natural language inference (NLI) with independently
translated propositional representations of the premise and hypothesis. Before
symbolic reasoning, the system must infer semantic links between their
proposition-level units (atoms). Existing systems based on
Abstract Meaning Representation (AMR) and Boolean satisfiability (SAT) identify
these links using fixed embedding-based source selection and pretrained
sentence-level NLI models not trained for atom-level linking.
We introduce a trainable atom linker that jointly predicts link existence,
source selection, and relation type.
Without atom-pair training annotations, we treat cross-formula link structures
as latent and train the linker from sentence-level NLI labels, keeping the
translator and SAT reasoner fixed.
We evaluate on the full public test splits of Stanford Natural Language Inference
(SNLI) and Sentences Involving Compositional Knowledge (SICK), and the matched
and mismatched validation splits of Multi-Genre Natural Language Inference
(MNLI). Our system improves accuracy by 11.40--27.09 percentage points over a
baseline applying sentence-trained NLI models unchanged to atom pairs
within the same SAT pipeline. It also improves over the prior fixed-linker
AMR-to-SAT system by 8.14--26.32 percentage points on SNLI and both MNLI splits,
with a smaller gain on SICK.
\end{abstract}

\section{Introduction}
\label{sec:introduction}

Natural language inference (NLI) classifies the relation between a premise
$P$ and a hypothesis $H$ as entailment, contradiction, or neutral
\citep{dagan-etal-2006-pascal,bowman-etal-2015-large,
williams-etal-2018-broad}.

We investigate whether learning links between independently translated
propositional formulae from sentence-level labels improves NLI accuracy with
the translator and inference procedure held fixed.

Logic-based NLI classifies sentence pairs by reasoning over formal representations
\citep{bos-markert-2005-recognising,yanaka-etal-2018-acquisition,
hu-etal-2020-monalog,abzianidze-2020-learning}.
Other work learns to score abductive assumptions
\citep{raina-etal-2005-robust} or incorporates distributional knowledge into
logical inference
\citep{beltagy-etal-2013-montague,beltagy-etal-2016-representing}.
Such systems expose their inference theory, but errors or omissions in
these representations can lead to incorrect conclusions. Neural NLI
predicts sentence relations directly from premise--hypothesis pairs,
typically without a formal derivation
\citep{rocktaschel-etal-2016-reasoning,parikh-etal-2016-decomposable}.

Neuro-symbolic NLI combines neural components with explicit inference
procedures \citep{kalouli-etal-2020-hy,chen-etal-2021-neurallog,
kobayashi-etal-2026-neural}.
Language models have been coupled with theorem provers
\citep{skandalis-etal-2025-neurosymbolic,noble-etal-2025-mood}.
Other systems jointly learn fact extraction and NLI
\citep{popovic-farber-2025-extractive}, or train logical translators using
entailment labels \citep{lee-etal-2025-entailment}.

Alignment models link premise and hypothesis structures
\citep{chambers-etal-2007-learning,maccartney-etal-2008-phrase}.
Sentence-level supervision has been used to learn latent alignments and
semantic relations \citep{watanabe-etal-2012-latent} and natural-logic
inference \citep{feng-etal-2020-exploring,feng-etal-2022-neuro}.
\citet{wu-etal-2023-weakly} align phrases by similarity and learn their
relations using fuzzy logic.
Logical aggregation has been used for span-level predictions
\citep{stacey-etal-2022-logical} and for relations between generated premise
facts and the full hypothesis \citep{stacey-etal-2024-atomic}.

Hypothesis decomposition has been used to diagnose NLI consistency
\citep{srikanth-rudinger-2025-nli} and construct Atomic-SNLI, which pairs full
premises with atomic hypotheses \citep{huang-2026-atomic}.

Prior systems independently translate the premise and hypothesis into
propositional formulae derived from Abstract Meaning Representation (AMR) and
use a Boolean satisfiability (SAT) solver to classify their relation
\citep{anonymous-ictai,anonymous-comma}.  We call each AMR-derived
propositional variable an \emph{atom} and associate it with a short
verbalization.  An \emph{atom link} connects a premise atom (source) to a
hypothesis atom (target).  Targets with exact verbalization matches are linked
deterministically; each remaining target may remain unlinked or receive a
source-specific entailment or contradiction link.

For example, let the premise formula be $\Phi_P=a\land b$, where $a$
verbalizes as \emph{dog runs} and $b$ as \emph{cat sleeps}, and let the
hypothesis formula be $\Phi_H=c$, where $c$ verbalizes as \emph{dog moves}.
The intended link is entailment from $a$ to $c$.
Atom linking therefore involves deciding whether to link each target, which
premise atom to select, and whether the relation is entailment or contradiction.

Prior atom-linking procedures use embedding similarity to select a source atom
and assign an entailment link when the selected pair is sufficiently similar;
otherwise, the target remains unlinked. A later system applies a pretrained
sentence-level NLI model to the selected pair to detect contradiction
\citep{anonymous-ictai,anonymous-comma}. In that system, the source selector
and NLI model are fixed. The NLI model is trained on complete sentence pairs
but applied unchanged to short atom verbalizations.

We replace these fixed components with a trainable atom linker that makes
these decisions jointly. Because our training data contain no atom-pair
annotations, we treat the link structure as latent and learn it from
sentence-level NLI labels. During training,
we use the SAT reasoner to label candidate link structures and train the
linker to favor structures that produce the observed sentence label.
At test time, the reasoner predicts the sentence relation using the link
structure assigned the highest probability by the linker.

Our \emph{Sentence-NLI Transfer} baseline applies a sentence-trained scorer
unchanged to candidate atom pairs within the same link-selection and SAT
pipeline as Learned Linker.

On the full public SNLI and SICK test splits and MNLI matched/mismatched
validation splits, the learned linker improves accuracy over Sentence-NLI
Transfer by 11.40--27.09 percentage points. Gains over the prior fixed-linker
AMR-to-SAT system are 8.14--26.32 points on SNLI/MNLI, with a smaller gain
on SICK. Balanced atom-relation probes evaluate the linker's
relation predictions; an error analysis examines limitations of the
learned links and fixed representation.

Code and reproduction materials for this work are available online.\footnote{\raggedright
\url{https://github.com/fxy-1117/Proposition-Level-Linker-for-Neuro-Symbolic-NLI}.}

\section{Background}
\label{sec:background}

Abstract Meaning Representation (AMR) encodes a sentence as a rooted, labeled,
directed graph \citep{banarescu-etal-2013-abstract}.  Nodes denote concepts or
PropBank frames.  Edges encode core roles such as \texttt{:ARG0} and
\texttt{:ARG1} \citep{kingsbury-palmer-2002-treebank}, as well as relations such
as \texttt{:location} and \texttt{:time}.  AMR can represent coordination,
conditions, polarity, and reentrancy.

AMR4NLI compares premise and hypothesis AMRs using asymmetric graph-matching measures
\citep{opitz-etal-2023-amr4nli}, while subsequent studies find that supplying
AMR as text or graph input does not consistently improve NLI models
\citep{raut-etal-2025-llms,min-etal-2025-meaning}.

AMR-to-logic mappings support symbolic and neuro-symbolic reasoning
\citep{bos-2016-squib,lai-etal-2020-continuation,chanin-hunter-2023-neuro}.
We use a fixed rule-based AMR-to-propositional translator
\citep{feng2026pairwise}, which maps each AMR graph to atoms for linking and
a propositional formula for SAT inference.\footnote{\raggedright Implementation and translation rules:
\url{https://github.com/fxy-1117/amr-to-propositional-logic}.}
Coordination yields conjunction or disjunction,
conditions yield implication, and formula-level polarity yields negation.

\begin{definition}[Statement-level propositional representation]
\label{def:statement-representation}
For a finite atom set $\mathcal{A}$, let $\mathcal{L}(\mathcal{A})$ be the
smallest formula language containing every $a\in\mathcal{A}$ and closed under
$\neg$, $\land$, $\lor$, and $\to$.  Each successfully translated statement
$x$ has the representation
$\mathsf{Rep}(x)=\langle\mathcal{A}_x,\Phi_x,\mathcal{V}_x\rangle$, where
$\mathcal{A}_x$ is a finite, nonempty set of active semantic atoms,
$\Phi_x\in\mathcal{L}(\mathcal{A}_x)$ is the formula, and
$\mathcal{V}_x$ maps each atom to a deterministic, nonempty verbalization.
Formula-level polarity occurs
in $\Phi_x$, not in $\mathcal{V}_x$.  Let $\operatorname{Atoms}(\Phi_x)$ denote
the atoms occurring in $\Phi_x$.  The translator satisfies
$\operatorname{Atoms}(\Phi_x)=\mathcal{A}_x$.
\end{definition}

The main atom forms are $\operatorname{Uni}(u)$,
$\operatorname{Dya}_{\kappa}(u,v)$, and
$\operatorname{Tri}_{\kappa_a,\kappa_b}(u,c,v)$. Here $u$, $c$, and $v$ are
AMR nodes and the subscripts identify roles. Unary atoms retain otherwise
unexpressed concepts; dyadic atoms encode one role relation. Event triples
join two roles at a predicate occurrence.

\begin{example}
\label{ex:amr-translation}
Consider $x^+$, \emph{A cute dog chases a cat}, and $x^-$,
\emph{No cute dog chases a cat}. Their AMRs differ only in root polarity:
\par\smallskip\noindent
\begingroup
\small
\begin{minipage}[t]{0.48\linewidth}
\vspace{0pt}
\centering
\textit{Positive AMR}\\
\raggedright
\texttt{(c / chase-01}\\
\quad\texttt{:ARG0 (d / dog}\\
\qquad\texttt{:mod (u / cute))}\\
\quad\texttt{:ARG1 (a / cat))}
\end{minipage}
\hfill
\begin{minipage}[t]{0.48\linewidth}
\vspace{0pt}
\centering
\textit{Negative AMR}\\
\raggedright
\texttt{(c / chase-01}\\
\quad\texttt{:polarity -}\\
\quad\texttt{:ARG0 (d / dog}\\
\qquad\texttt{:mod (u / cute))}\\
\quad\texttt{:ARG1 (a / cat))}
\end{minipage}
\endgroup

\par\smallskip\noindent
The translator constructs
$a_1=\operatorname{Dya}_{\mathrm{mod}}(\textit{dog},\textit{cute})$ and
$a_2=\operatorname{Tri}_{\mathrm{ARG0},\mathrm{ARG1}}
(\textit{dog},\textit{chase-01},\textit{cat})$.
Both statements have atom set $\{a_1,a_2\}$ and verbalizations
\emph{cute dog} and \emph{dog chase cat}.  The formulae are
$\Phi_{x^+}=a_2\land a_1$ and
$\Phi_{x^-}=\neg(a_2\land a_1)$.
\end{example}

\section{Method}
\label{sec:method}

\begin{figure*}[t]
\centering
\begin{tikzpicture}[
  >=Latex,
  sentence/.style={draw=gray!55, fill=gray!5, rounded corners=2pt,
    text width=7.05cm, align=left, inner sep=5pt,
    minimum height=1.30cm, font=\small},
  representation/.style={draw=gray!55, rounded corners=2pt,
    minimum width=7.40cm, minimum height=2.65cm},
  atom/.style={draw=gray!40, fill=gray!3, rounded corners=1.5pt,
    text width=5.70cm, align=left, inner sep=3pt, font=\footnotesize},
  linker/.style={draw=blue!40!black, fill=blue!4, rounded corners=2pt,
    text width=15.30cm, align=center, inner sep=5pt, font=\small},
  panel/.style={draw=gray!60, fill=gray!2, rounded corners=2pt,
    minimum width=7.40cm, minimum height=6.95cm},
  heading/.style={font=\small\bfseries, text=blue!40!black},
  paneltext/.style={text width=6.85cm, align=left, font=\small,
    inner sep=0pt},
  flow/.style={-{Latex[length=2mm]}, thick, draw=gray!65},
  update/.style={-{Latex[length=2mm]}, thick, dashed, draw=blue!60!black},
  entlink/.style={-{Latex[length=2mm]}, very thick, draw=green!40!black},
  conlink/.style={-{Latex[length=2mm]}, very thick, draw=red!70!black}
]

\node[sentence] (psentence) at (-4.00,7.40) {\textbf{Premise $P$}\\
``Two men are looking through microscopes in a room.''};
\node[sentence] (hsentence) at (4.00,7.40) {\textbf{Hypothesis $H$}\\
``The men are watching TV in a room.''};

\node[representation] (premisebox) at (-4.00,4.925) {};
\node[representation] (hypothesisbox) at (4.00,4.925) {};
\draw[flow] (psentence.south) -- (premisebox.north);
\draw[flow] (hsentence.south) -- (hypothesisbox.north);
\node[font=\small] at (0,6.50) {Fixed AMR parsing and propositional translation};

\node[heading] at (-4.00,5.92) {Premise representation};
\node[heading] at (4.00,5.92) {Hypothesis representation};
\node[font=\small] at (-4.00,5.49)
  {$\mathcal{A}_P=\{s_1,s_2\}$,\quad $\Phi_P=s_1\land s_2$};
\node[font=\small] at (4.00,5.49)
  {$\mathcal{A}_H=\{t_1,t_2\}$,\quad $\Phi_H=t_1\land t_2$};
\node[atom] (p1) at (-4.00,4.98)
  {$s_1$: \texttt{2 men look microscope}};
\node[atom] (p2) at (-4.00,4.38)
  {$s_2$: \texttt{2 men look at room}};
\node[atom] (h1) at (4.00,4.98)
  {$t_1$: \texttt{man watch television}};
\node[atom] (h2) at (4.00,4.38)
  {$t_2$: \texttt{man watch at room}};
\node[font=\footnotesize] at (4.00,3.85)
  {$\mathcal{U}=\{t_1,t_2\}$; no exact verbalization matches};

\node[linker] (linker) at (0,2.90)
  {\textbf{Trainable atom linker}\quad Target-wise action distributions};
\draw[flow] (premisebox.south) -- (premisebox.south |- linker.north);
\draw[flow] (hypothesisbox.south) -- (hypothesisbox.south |- linker.north);

\node[panel] (training) at (-4.00,-1.375) {};
\node[panel] (inference) at (4.00,-1.375) {};
\draw[flow] (training.north |- linker.south) -- (training.north);
\draw[flow] (inference.north |- linker.south) -- (inference.north);
\node[heading] at (-4.00,1.82) {(a) Training: label-specific witnesses};
\node[heading] at (4.00,1.82) {(b) Inference: one MAP world};

\node[paneltext, anchor=north] at (-4.00,1.50)
  {\textbf{1. Epoch-start MAP:} $\bar w_e=(\mathsf{C}_{11},\mathsf{E}_{22})$\\[2pt]
Hamming-1 pool: $9$ of $25$ worlds ($3$ shown).};
\node[anchor=north, inner sep=0pt, font=\small] (pool) at (-4.00,0.57) {
  \begin{tabular}{@{}lcc@{}}
  \toprule
  World $w$ & Targets changed & SAT label \\
  \midrule
  $(\mathsf{C}_{11},\mathsf{E}_{22})$ & $0$ & $\Con$ \\
  $(\varnothing,\mathsf{E}_{22})$ & $1$ & $\Neu$ \\
  $(\mathsf{E}_{11},\mathsf{E}_{22})$ & $1$ & $\Ent$ \\
  \bottomrule
  \end{tabular}
};
\node[paneltext, anchor=north] at (-4.00,-1.42)
  {Pool and SAT labels are fixed within the epoch.};
\node[paneltext, anchor=north] (witnesses) at (-4.00,-2.00)
  {\textbf{2. Select label-specific witnesses}\\[2pt]
For each reachable SAT label, select the highest-scoring world with that
label from all $9$, using \mbox{current $\theta$}.};
\node[paneltext, anchor=north] (loss) at (-4.00,-3.82)
  {\textbf{3. Cross-entropy over witness scores}\\[2pt]
Gold sentence label: $\Con$.};
\coordinate (updateaxis) at (-0.13,0);
\draw[update] (loss.east) -- (loss.east -| updateaxis) --
  (updateaxis |- linker.south);
\node[font=\footnotesize, text=blue!60!black, rotate=90, inner sep=0pt]
  at (0.13,-0.30) {Update $\theta$};

\node[paneltext, anchor=north] at (4.00,1.50) {\textbf{1. MAP link world:}
$\hat w=(\mathsf{C}_{11},\mathsf{E}_{22})$.};
\node[font=\small] (map-p1) at (1.15,0.70) {$s_1$};
\node[font=\small] (map-h1) at (3.35,0.70) {$t_1$};
\node[font=\small] (map-p2) at (4.65,0.70) {$s_2$};
\node[font=\small] (map-h2) at (6.85,0.70) {$t_2$};
\draw[conlink] (map-p1.east) -- node[above, inner sep=1pt,
  font=\footnotesize, text=red!70!black] {$\mathsf{C}_{11}$} (map-h1.west);
\draw[entlink] (map-p2.east) -- node[above, inner sep=1pt,
  font=\footnotesize, text=green!35!black] {$\mathsf{E}_{22}$} (map-h2.west);
\node[paneltext, anchor=north] (worldformula) at (4.00,-0.60)
  {\textbf{2. World formula}\\[3pt]
$\Phi_{\hat w}=\Phi_P\land
(\neg s_1\lor\neg t_1)\land(\neg s_2\lor t_2)$.};
\node[paneltext, anchor=north] (satprediction) at (4.00,-2.25)
  {\textbf{3. SAT checks}\\[3pt]
$\Phi_{\hat w}$ satisfiable;\\[2pt]
$\Phi_{\hat w}\land\Phi_H$ unsatisfiable.\\[4pt]
\textbf{Prediction:} $\hat y=\Con$.};

\end{tikzpicture}
\caption{A held-out SNLI example (gold $\Con$) with independently translated
formulae and atom texts. (a) Training is illustrated using the MAP structure
in (b) as the pool center; this example is not used for training.
(b) Colored arrows show the inference MAP links, which yield SAT label $\Con$.}
\label{fig:snli-interpretability}
\end{figure*}
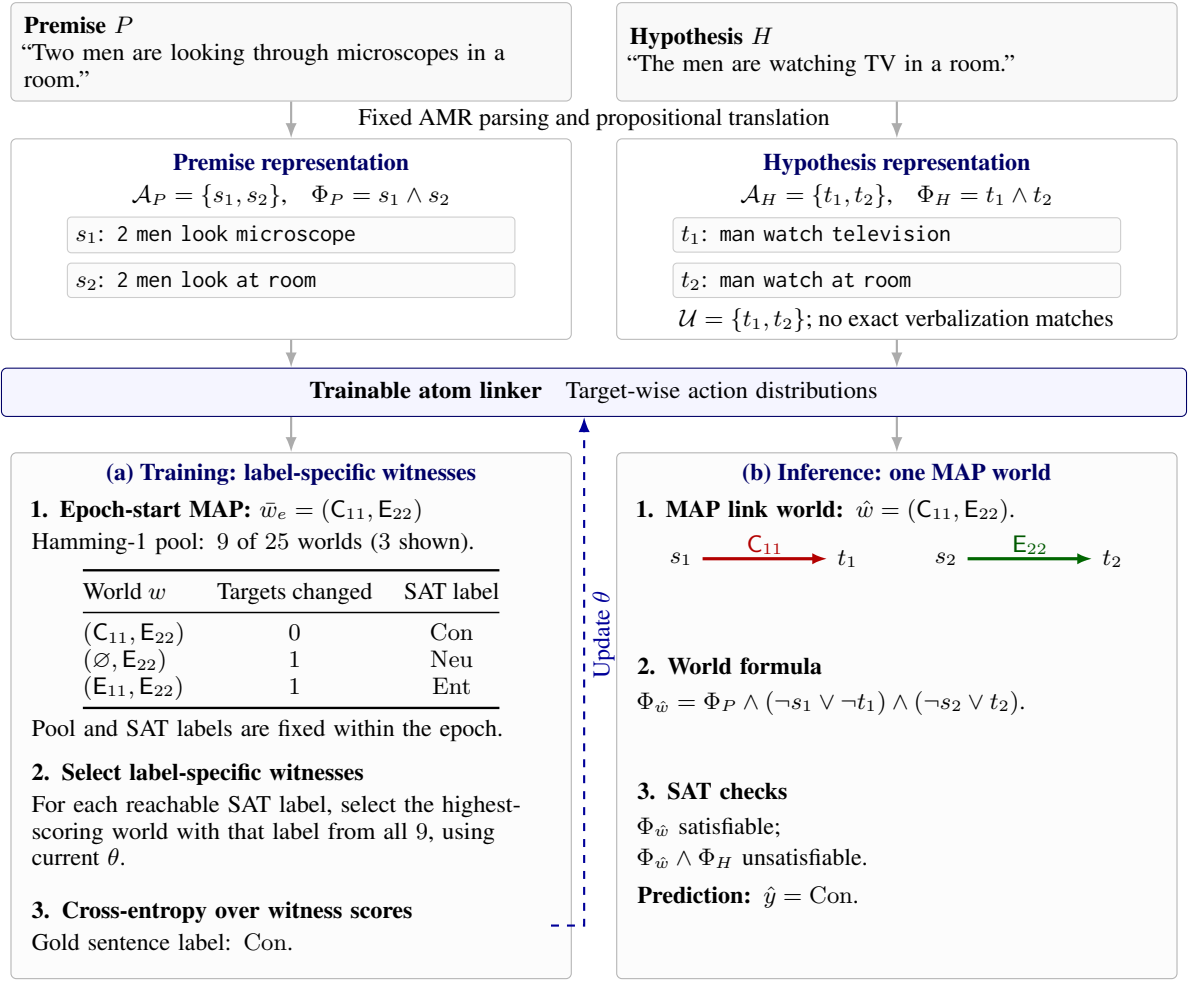

\paragraph{Overview.}
A fixed AMR parser and translator map the premise $P$ and hypothesis $H$ to
propositional representations.
Hypothesis atoms with deterministic exact matches bypass the neural linker.
For each remaining hypothesis atom, a trainable cross-encoder scores every
premise atom and defines a categorical choice among no
link and source-specific entailment or contradiction links. At test time,
target-wise maximum a posteriori (MAP) choices select one link world.
The SAT reasoner combines its link clauses and the exact-match clauses with
the premise formula, then checks the hypothesis to assign an NLI label. During
training, sentence labels supervise the linker through an epoch-start MAP link
structure and alternatives obtained by changing one target's action.
Figure~\ref{fig:snli-interpretability} illustrates training and inference.

\subsection{Fixed Propositional Representations}
\label{sec:fixed-representations}

We rename premise and hypothesis atoms into disjoint namespaces so that
identical names do not equate unrelated propositions. The formulae and
verbalization maps are renamed consistently.

\begin{definition}[Premise and hypothesis representation]
\label{def:nli-representation}
After renaming, Definition~\ref{def:statement-representation} gives
$\mathsf{Rep}(X)=\langle\mathcal{A}_X,\Phi_X,\mathcal{V}_X\rangle$ for
$X\in\{P,H\}$, where
$\mathcal{A}_P\cap\mathcal{A}_H=\varnothing$.
\end{definition}

\begin{definition}[Exact matching and candidates]
\label{def:candidate-atom-links}
Define the exact-match set and unmatched hypothesis atoms by
\begin{equation}
\begin{aligned}
\mathcal{M}^{\mathrm{ex}}
&=\{(a,b)\in\mathcal{A}_P\times\mathcal{A}_H:\\
&\qquad\mathcal{V}_P(a)=\mathcal{V}_H(b)\},\\
\mathcal{U}
&=\{b\in\mathcal{A}_H:\nexists a\in\mathcal{A}_P,
(a,b)\in\mathcal{M}^{\mathrm{ex}}\}.
\end{aligned}
\label{eq:exact-and-unmatched}
\end{equation}
All exact matches are retained. Exactly matched targets bypass neural
scoring, while their source atoms remain available to unmatched targets.

Index the premise atoms as $\mathcal{A}_P=\{s_i\}_{i=1}^{n}$ in fixed
atom-identifier order, and write $\mathcal{U}=\{t_j\}_{j=1}^{m}$.
Every premise atom is a candidate source for each unmatched hypothesis atom.
\end{definition}

\begin{example}
\label{ex:candidate-atom-links}
Let $s_1$ and $s_2$ verbalize as \emph{dog exists} and \emph{dog runs},
and let $t_0$ and $t_1$ verbalize as \emph{dog exists} and
\emph{dog moves}.  Then
$\mathcal{M}^{\mathrm{ex}}=\{(s_1,t_0)\}$,
$\mathcal{U}=\{t_1\}$.
Although $s_1$ exactly matches $t_0$, it remains a candidate source for $t_1$.
\end{example}

\subsection{Target-Wise Atom-Link Distribution}
\label{sec:neural-linker}

\begin{definition}[Atom-pair natural language inference]
\label{def:atom-nli}
Let $\mathsf{N}_\theta$ be a trainable cross-encoder that jointly encodes
nonempty strings $u$ (premise) and $v$ (hypothesis) and returns
\[
\mathsf{N}_\theta(u,v)=\bigl(\ell^E_\theta(u,v),\ell^C_\theta(u,v),
\ell^N_\theta(u,v)\bigr)\in\mathbb{R}^{3}.
\]
The logits correspond to entailment,
contradiction, and neutral, respectively; $\theta$ denotes the trainable
linker parameters.
\end{definition}

To jointly choose a source and relation, we normalize the link actions and
a no-link action in one distribution per target.

\begin{definition}[Target-wise link distribution]
\label{def:world-distribution}
For every $s_i\in\mathcal{A}_P$, let $\mathsf{E}_{ij}$ and
$\mathsf{C}_{ij}$ denote the entailment and contradiction link actions from
$s_i$ to $t_j$. The ordered legal actions for $t_j$ are
\begin{equation}
\mathcal{Z}_j
=\bigl(\varnothing,
\mathsf{C}_{1j},\mathsf{E}_{1j},\ldots,
\mathsf{C}_{nj},\mathsf{E}_{nj}\bigr),
\label{eq:legal-actions}
\end{equation}
where $\varnothing$ denotes no link. Define the pair logits by
\[
(\ell^E_{ij},\ell^C_{ij},\ell^N_{ij})
=\mathsf{N}_\theta\bigl(\mathcal{V}_P(s_i),\mathcal{V}_H(t_j)\bigr).
\]
The no-link action represents the all-neutral configuration: no candidate
source receives a non-neutral relation to $t_j$. We score each link action
relative to this reference:
\begin{align}
A_{\theta,j}(\varnothing)&=0,\quad
A_{\theta,j}(\mathsf{E}_{ij})=\ell^E_{ij}-\ell^N_{ij},\nonumber\\
A_{\theta,j}(\mathsf{C}_{ij})&=\ell^C_{ij}-\ell^N_{ij},
\label{eq:choice-scores}\\
q_{\theta,j}(z\mid P,H)
&=\frac{\exp A_{\theta,j}(z)}
{\sum_{z'\in\mathcal{Z}_j}\exp A_{\theta,j}(z')}.
\label{eq:local-softmax}
\end{align}

Here, $q_{\theta,j}(z\mid P,H)$ is the probability of choosing action $z$
for target $t_j$.
A link world assigns one legal action to every unmatched hypothesis atom:
$\mathcal{W}=\prod_{j=1}^{m}\mathcal{Z}_j$, with
$w=(z_1,\ldots,z_m)\in\mathcal{W}$.
Assuming conditional independence across target-level choices given $P,H$,
the probability of a complete link world is the product of its action
probabilities:
\begin{equation}
p_\theta(w\mid P,H)
=\prod_{j=1}^{m}q_{\theta,j}(z_j\mid P,H).
\label{eq:world-distribution-score}
\end{equation}
If $m=0$, $\mathcal{W}=\{()\}$ and $p_\theta(()\mid P,H)=1$.
\end{definition}

\begin{example}
\label{ex:action-world-probability}
Consider one premise atom and two unmatched targets. For $(s_1,t_1)$,
classifier logits $(\ell^E_{11},\ell^C_{11},\ell^N_{11})=(\log 8,0,0)$
give the following action scores and probabilities:
\begin{center}
\small
\begin{tabular}{@{}lcc@{}}
\toprule
Action $z$ & Score & Probability \\
\midrule
$\varnothing$ & $0$ & $0.1$ \\
$\mathsf{C}_{11}$ & $\ell^C_{11}-\ell^N_{11}=0$ & $0.1$ \\
$\mathsf{E}_{11}$ & $\ell^E_{11}-\ell^N_{11}=\log 8$ & $0.8$ \\
\bottomrule
\end{tabular}
\end{center}
For $t_2$, let the probabilities in action order
$(\varnothing,\mathsf{C}_{12},\mathsf{E}_{12})$ be $(0.2,0.1,0.7)$.
Then world
$w=(\mathsf{E}_{11},\varnothing)$ has probability
$p_\theta(w\mid P,H)=0.8\times0.2=0.16$.
\end{example}

\subsection{Single-MAP Logical Inference}
\label{sec:inference}

Test-time inference selects one complete link structure from the world
distribution.

\begin{definition}[Single-MAP link world]
\label{def:single-map-world}
For each unmatched target, choose
$\hat z_j\in\argmax_{z\in\mathcal{Z}_j}q_{\theta,j}(z\mid P,H)$ and set
$\hat w=(\hat z_1,\ldots,\hat z_m)$.
Maximizing each target-wise probability maximizes their product, so
$\hat w\in\argmax_{w\in\mathcal{W}}p_\theta(w\mid P,H)$.
Ties are resolved by selecting the first maximizing action in the order of
Equation~\ref{eq:legal-actions}.
\end{definition}

\begin{example}
\label{ex:map-world-probability}
Continuing Example~\ref{ex:action-world-probability}, selecting each target's
most probable action gives $\hat w=(\mathsf{E}_{11},\mathsf{E}_{12})$,
with probability $0.8\times0.7=0.56$.
\end{example}

The selected world determines the clauses conjoined with the premise formula.

\begin{definition}[Hard link clauses and world formula]
\label{def:world-theory}
Each deterministic exact verbalization match contributes a hard equivalence:
\begin{equation}
\Omega^{\mathrm{ex}}
=\!\bigcup_{(a,b)\in\mathcal{M}^{\mathrm{ex}}}
\{[\neg a\lor b],[a\lor\neg b]\}.
\label{eq:exact-clauses}
\end{equation}
For a neural link action, define
$\operatorname{cl}(\mathsf{E}_{ij})=[\neg s_i\lor t_j]$ and
$\operatorname{cl}(\mathsf{C}_{ij})=[\neg s_i\lor\neg t_j]$.
An entailment action contributes the directional implication
$s_i\to t_j$.  A contradiction action contributes the incompatibility
$\neg(s_i\land t_j)$.

For $w=(z_1,\ldots,z_m)$, the world formula is
\begin{equation}
\Phi_w=\Phi_P\land
\bigwedge_{\gamma\in\Omega^{\mathrm{ex}}}\gamma
\land\bigwedge_{\substack{1\leq j\leq m\\z_j\neq\varnothing}}
\operatorname{cl}(z_j).
\label{eq:world-theory}
\end{equation}
Empty conjunctions are true.
\end{definition}

\begin{definition}[World-level NLI label]
\label{def:world-label}
Let $\mathcal{Y}=\{\Ent,\Con,\Neu,\Inv\}$. Write
$\operatorname{UNSAT}(\psi)$ when a formula $\psi$ has no satisfying
assignment. Define
\begin{equation}
\Lworld(w)=
\begin{cases}
\Inv, & \operatorname{UNSAT}(\Phi_w),\\
\Con, & \operatorname{UNSAT}(\Phi_w\land\Phi_H),\\
\Ent, & \operatorname{UNSAT}(\Phi_w\land\neg\Phi_H),\\
\Neu, & \text{otherwise}.
\end{cases}
\label{eq:world-label}
\end{equation}
The first applicable case determines the label, so inconsistent worlds
receive $\Inv$. For a consistent world, $\Con$ means that it entails
$\neg\Phi_H$, $\Ent$ means it entails $\Phi_H$, and $\Neu$ means it entails
neither.
\end{definition}

Appendix~\ref{app:link-construction} describes the SAT implementation.
Evaluating only the MAP world gives the final prediction
$\hat y=\Lworld(\hat w)$.

\begin{example}
\label{ex:fixed-world-inference}
Using Example~\ref{ex:candidate-atom-links}, let
$\Phi_P=s_1\land s_2$ and $\Phi_H=t_0\land t_1$. Suppose the
probabilities over
$\mathcal{Z}_1=(\varnothing,\mathsf{C}_{11},\mathsf{E}_{11},
\mathsf{C}_{21},\mathsf{E}_{21})$ are
$(0.10,0.05,0.05,0.10,0.70)$. Definition~\ref{def:single-map-world} selects
$\hat w=(\mathsf{E}_{21})$.  The exact match adds
$[\neg s_1\lor t_0]$ and $[s_1\lor\neg t_0]$, while
$\mathsf{E}_{21}$ adds $[\neg s_2\lor t_1]$; hence
$\Phi_{\hat w}\models\Phi_H$ and $\hat y=\Ent$. Selecting
$\mathsf{C}_{21}$ instead yields $\Con$, whereas selecting $\varnothing$
leaves $t_1$ unconstrained and yields $\Neu$.
\end{example}

\subsection{Label-Specific Latent-Witness Training}
\label{sec:latent-witness-training}

Sentence labels do not specify link structures. The number of worlds grows
exponentially with the number of unmatched targets. Training therefore uses
the epoch-start MAP world and all worlds that differ from it in one target
action. These alternative link structures can receive different SAT labels,
allowing the training objective to compare label-specific witnesses.

\begin{definition}[Epoch-start MAP-centered Hamming-1 pool]
\label{def:hamming-pool}
Let $\bar\theta_e$ denote the linker parameters at the start of epoch $e$.
We construct the pool using $\bar\theta_e$ with dropout disabled. Define
\[
\begin{gathered}
\bar z_{e,j}\in\argmax_{z\in\mathcal{Z}_j}q_{\bar\theta_e,j}(z\mid P,H),\\
\bar w_e=(\bar z_{e,1},\ldots,\bar z_{e,m}).
\end{gathered}
\]
Ties are resolved as in Definition~\ref{def:single-map-world}.
Let $d_H(w,w')$ be the number of targets whose actions differ between
$w$ and $w'$. The pool contains $\bar w_e$ and every world obtained by
replacing exactly one of its target actions with another legal action:
\begin{equation}
\Wlocal(\bar w_e)
=\{\bar w_e\}\cup\{w\in\mathcal{W}:d_H(w,\bar w_e)=1\}.
\label{eq:hamming-one}
\end{equation}
Its size is $|\Wlocal(\bar w_e)|=1+2nm$, compared with
$|\mathcal{W}|=(2n+1)^m$ for the full world space.
\end{definition}

The pool and its SAT labels remain fixed throughout the epoch; world scores
and label-specific witnesses are recomputed using the current parameters
$\theta$.

\begin{example}
\label{ex:hamming-pool}
Let $\Phi_P=s_1$ and $\Phi_H=t_1\land t_2$, with one premise atom
and both hypothesis atoms unmatched. Each target has three actions, ordered
as $(\varnothing,\mathsf{C},\mathsf{E})$, with epoch-start distributions
$q_{\bar\theta_e,1}=(0.10,0.10,0.80)$ and
$q_{\bar\theta_e,2}=(0.50,0.20,0.30)$.
Then $\bar w_e=(\mathsf{E}_{11},\varnothing)$ and the complete Hamming-1
pool contains five of the nine possible worlds, with reasoner labels shown in the final column:
\begin{center}
\small
\begin{tabular}{lcc}
\toprule
World $w$ & $p_{\bar\theta_e}(w\mid P,H)$ & $\Lworld(w)$ \\
\midrule
$(\mathsf{E}_{11},\varnothing)$ & $0.40$ & $\Neu$ \\
$(\varnothing,\varnothing)$ & $0.05$ & $\Neu$ \\
$(\mathsf{C}_{11},\varnothing)$ & $0.05$ & $\Con$ \\
$(\mathsf{E}_{11},\mathsf{C}_{12})$ & $0.16$ & $\Con$ \\
$(\mathsf{E}_{11},\mathsf{E}_{12})$ & $0.24$ & $\Ent$ \\
\bottomrule
\end{tabular}
\end{center}
These are probabilities under the full world distribution, so they need not
sum to one within the pool.
\end{example}

A label is reachable in the pool if the SAT reasoner assigns it to at
least one world in the pool. For each reachable label, we use its
highest-scoring world as a witness and apply cross-entropy to the
witness scores.

\begin{definition}[Label-specific latent-witness objective]
\label{def:latent-witness-objective}
For each $y\in\mathcal{Y}$, define the epoch-$e$ label-specific world set and
witness score
\begin{align}
\mathcal{W}_{e,y}
&=\{w\in\Wlocal(\bar w_e):\Lworld(w)=y\},\nonumber\\
g_{\theta,e}(y)
&=\max_{w\in\mathcal{W}_{e,y}}\log p_\theta(w\mid P,H).
\label{eq:label-witness}
\end{align}
We use the convention $\max\varnothing=-\infty$.
For an observed label $y^*\in\{\Ent,\Con,\Neu\}$ that is reachable in the
local pool, the per-example loss is
\begin{equation}
\mathcal{L}_{\mathrm{LW}}
=-g_{\theta,e}(y^*)
+\log\sum_{y\in\mathcal{Y}}\exp g_{\theta,e}(y).
\label{eq:latent-witness-loss}
\end{equation}
Neural links leave the satisfiability of the premise and exact-match base
unchanged: each unmatched target occurs in at most one added clause.
Thus $\Inv$ cannot compete with a reachable gold label.

Let $\mathcal{D}$ index the training examples and define
\begin{align}
\mathcal{D}_e^+
&=\{i\in\mathcal{D}:\mathcal{W}^{(i)}_{e,y_i^*}\neq\varnothing\},\nonumber\\
\mathcal{L}_e(\theta)
&=\frac{1}{|\mathcal{D}|}
\sum_{i\in\mathcal{D}_e^+}\mathcal{L}^{(i)}_{\mathrm{LW}}.
\label{eq:training-objective}
\end{align}
Gold-unreachable examples contribute zero but remain in the normalization.
\end{definition}

\begin{example}
\label{ex:latent-witness-loss}
Continuing Example~\ref{ex:hamming-pool}, at $\theta=\bar\theta_e$ the highest
world probabilities for $\Ent$, $\Neu$, and $\Con$ are $0.24$, $0.40$, and
$0.16$, respectively. If the gold label is $\Ent$, then
$\mathcal{L}_{\mathrm{LW}}=-\log 0.24+
\log(0.24+0.40+0.16)\approx1.204$.
The $\Ent$ witness replaces the MAP world's no-link action for $t_2$ with
$\mathsf{E}_{12}$; the loss favors it over the competing label witnesses.
\end{example}

A MAP-only pool provides no gradient from this objective: the loss is zero
when its label matches the gold label, and the example contributes zero
otherwise. Examples with $m=0$ likewise provide no linker gradient.

\section{Experiments}
\label{sec:experiments}

\subsection{Datasets}

We evaluate on three English three-class NLI benchmarks: Stanford Natural Language
Inference (SNLI; \citealp{bowman-etal-2015-large}), Multi-Genre Natural Language
Inference (MNLI; \citealp{williams-etal-2018-broad}), and Sentences Involving
Compositional Knowledge (SICK; \citealp{marelli-etal-2014-semeval}). Evaluation
uses the full SNLI test split with valid labels (9,824 examples), the MNLI
matched (MNLI-m; 9,815) and mismatched (MNLI-mm; 9,832) validation splits,
and the 4,906-example public SICK test split.

For SNLI and MNLI, we randomly select 30,000 training examples per dataset,
with 10,000 per label; for SICK, we use 3,722 examples. Direct Sentence and
Learned Linker share these training sets. Sentence-NLI Transfer reuses the
corresponding Direct Sentence checkpoint without additional training.
All systems use the same development examples: 3,000 from SNLI validation,
3,000 held out from MNLI training, and all 495 SICK validation examples.
Appendix~\ref{app:data-selection} gives the selection details and SICK release.

\subsection{Systems and Evaluation}

We compare four systems. \emph{Direct Sentence MiniLM} is trained and evaluated
on complete premise--hypothesis pairs, providing a neural baseline without
propositional translation or SAT inference. \emph{Sentence-NLI Transfer} applies
the corresponding Direct Sentence checkpoint unchanged to atom pairs, then
uses our legal-choice space and single-MAP SAT reasoner, with no atom-level tuning.
\emph{Learned Atom Linker + SAT} shares Transfer's MiniLM architecture, fixed
atom representations, action space, and SAT reasoner, and trains the scorer
with the latent objective in Section~\ref{sec:latent-witness-training}.
Our main comparison is Learned Linker versus Sentence-NLI Transfer.
We evaluate \emph{FH-SAT} on the same propositional representations,
using its original fixed linker and inference procedure
\citep{anonymous-ictai,anonymous-comma}.

All neural systems use seeds 2026, 2027, and 2028. FH-SAT is deterministic,
with development-selected thresholds. MNLI-m and MNLI-mm share one checkpoint
per neural system and seed, and one FH-SAT threshold configuration.
We report Accuracy,
Macro-F1, class recall, and Invalid rate. Accuracy counts Invalid predictions
as errors over the full evaluation split. Macro-F1 averages F1 over
Entailment, Contradiction, and Neutral; Invalid predictions count as false
negatives for their gold class.

\begin{table*}[t]
\centering
\footnotesize
\setlength{\tabcolsep}{2.0pt}
\begin{tabular}{llr@{\hspace{8pt}}rrrrr}
\toprule
Data & System & Acc.$\uparrow$ & M-F1$\uparrow$ & $R_E\uparrow$ & $R_C\uparrow$ & $R_N\uparrow$ & Inv.$\downarrow$ \\
\midrule
SNLI & Direct Sentence & $85.23\pm0.36$ & $85.14\pm0.41$ & $87.72\pm1.22$ & $88.91\pm0.45$ & $78.94\pm2.54$ & 0.00 \\
 & Sentence-NLI Transfer & $51.13\pm2.26$ & $45.08\pm2.76$ & $48.39\pm5.43$ & $95.45\pm0.71$ & $9.42\pm1.97$ & 0.00 \\
 & FH-SAT & 51.90 & 48.65 & 74.67 & 16.19 & 64.00 & 1.85 \\
 & Learned Linker & $\mathbf{78.22\pm0.30}$ & $\mathbf{78.17\pm0.28}$ & $82.41\pm1.59$ & $79.13\pm0.72$ & $72.92\pm0.42$ & 0.00 \\
\addlinespace
MNLI-m & Direct Sentence & $79.11\pm0.45$ & $79.03\pm0.41$ & $81.21\pm1.83$ & $78.14\pm0.84$ & $77.76\pm1.66$ & 0.00 \\
 & Sentence-NLI Transfer & $39.41\pm1.04$ & $37.62\pm1.34$ & $28.60\pm1.40$ & $66.11\pm0.55$ & $23.98\pm2.88$ & 0.06 \\
 & FH-SAT & 53.26 & 54.85 & 65.36 & 44.91 & 48.35 & 7.14 \\
 & Learned Linker & $\mathbf{61.40\pm0.60}$ & $\mathbf{61.61\pm0.63}$ & $60.79\pm0.50$ & $58.04\pm2.31$ & $65.52\pm3.65$ & 0.06 \\
\addlinespace
MNLI-mm & Direct Sentence & $80.27\pm0.23$ & $80.18\pm0.20$ & $83.54\pm1.78$ & $78.97\pm0.49$ & $78.00\pm1.84$ & 0.00 \\
 & Sentence-NLI Transfer & $39.09\pm1.45$ & $37.13\pm1.87$ & $29.38\pm1.67$ & $66.05\pm0.54$ & $21.92\pm3.18$ & 0.10 \\
 & FH-SAT & 53.49 & 55.22 & 67.77 & 45.19 & 46.28 & 8.10 \\
 & Learned Linker & $\mathbf{62.00\pm0.39}$ & $\mathbf{62.17\pm0.42}$ & $63.35\pm0.78$ & $59.60\pm2.88$ & $63.00\pm3.48$ & 0.10 \\
\addlinespace
SICK & Direct Sentence & $87.10\pm0.37$ & $86.31\pm0.24$ & $86.02\pm2.12$ & $84.36\pm1.00$ & $88.34\pm1.84$ & 0.00 \\
 & Sentence-NLI Transfer & $72.77\pm9.78$ & $68.64\pm10.12$ & $56.24\pm11.64$ & $70.93\pm1.29$ & $81.55\pm11.78$ & 0.02 \\
 & FH-SAT & 82.10 & 80.55 & 73.29 & 73.46 & 88.75 & 0.00 \\
 & Learned Linker & $\mathbf{84.17\pm0.24}$ & $\mathbf{82.30\pm0.19}$ & $79.11\pm0.89$ & $70.79\pm0.61$ & $90.13\pm0.14$ & 0.02 \\
\bottomrule
\end{tabular}
\caption{Results on full public splits (\%). Neural values are three-seed mean $\pm$ sample standard deviation; FH-SAT is deterministic. Bold marks the highest Accuracy and Macro-F1 among Transfer, FH-SAT, and Learned Linker. $R_E/R_C/R_N$: class recalls; Inv.: Invalid rate.}
\label{tab:main-results}
\end{table*}

\subsection{Optimization and Model Selection}
\label{sec:optimization}

Separate SNLI, MNLI, and SICK linkers start from the pretrained
33M-parameter MiniLMv1 encoder \citep{wang-etal-2020-minilm}
and a newly initialized classification head. Linker training uses no
sentence-NLI checkpoint or atom-pair labels. We train the encoder and
classification head of Direct Sentence and Learned Linker with AdamW
\citep{loshchilov-hutter-2019-decoupled}.
Dataset-specific settings are reported in Appendix~\ref{app:hyperparameters}.

We select checkpoints by development Macro-F1, breaking ties by Accuracy.
Direct Sentence is evaluated on sentence pairs, and Learned Linker through
single-MAP SAT inference.
Evaluation splits are not used for checkpoint selection.

\section{Results}
\label{sec:results}

\subsection{Main Results}

As shown in Table~\ref{tab:main-results}, Learned Linker improves Accuracy over
Sentence-NLI Transfer on all four full splits by 11.40--27.09 percentage points;
Macro-F1 gains range from 13.65 to 33.09 points. The 95\% crossed paired-bootstrap
intervals exclude zero for both metrics. Accuracy gains are significant for
every seed under exact McNemar tests with Holm correction across the four
splits within each seed (Appendix~\ref{app:paired}). With representations,
legal actions, and SAT-based inference held fixed, these gains show that learning
atom links from sentence labels yields more accurate sentence predictions
than applying a sentence-trained scorer unchanged.

Compared with FH-SAT, Learned Linker improves Accuracy by
8.14--26.32 points on SNLI and both MNLI splits. Gains on SICK are smaller:
$+2.07$ Accuracy and $+1.75$ Macro-F1 points. The three-seed aggregate
95\% intervals are $[1.06,3.06]$ and $[0.64,2.86]$, respectively.
Direct Sentence remains more accurate on every split.

Appendices~\ref{app:scaling}--\ref{app:efficiency} report analyses of training
data size, reachable labels, and computational cost.

\subsection{Atom-Relation Probe}
\label{sec:atom-relation-probes}
\begin{table*}[t]
\centering
\footnotesize
\setlength{\tabcolsep}{2.5pt}
\begin{tabular}{llrrrrr}
\toprule
Data & System & Acc.$\uparrow$ & M-F1$\uparrow$ & $R_E\uparrow$ & $R_C\uparrow$ & $R_N\uparrow$ \\
\midrule
SNLI & Sentence-NLI Transfer & $64.52\pm1.41$ & $58.88\pm1.84$ & $97.33\pm0.88$ & $77.33\pm3.84$ & $18.89\pm2.71$ \\
 & Learned Linker & $\mathbf{74.26\pm1.94}$ & $\mathbf{72.92\pm1.89}$ & $98.67\pm0.33$ & $78.78\pm5.32$ & $45.33\pm1.67$ \\
\addlinespace
MNLI-m & Sentence-NLI Transfer & $66.44\pm0.22$ & $62.68\pm0.61$ & $85.89\pm1.90$ & $86.67\pm3.21$ & $26.78\pm2.78$ \\
 & Learned Linker & $\mathbf{67.78\pm1.20}$ & $\mathbf{64.59\pm1.33}$ & $97.11\pm0.69$ & $76.89\pm2.50$ & $29.33\pm2.03$ \\
\addlinespace
MNLI-mm & Sentence-NLI Transfer & $68.44\pm0.78$ & $64.49\pm1.69$ & $86.89\pm1.39$ & $91.11\pm1.64$ & $27.33\pm4.58$ \\
 & Learned Linker & $\mathbf{68.81\pm1.12}$ & $\mathbf{65.50\pm0.98}$ & $96.67\pm0.00$ & $79.78\pm3.53$ & $30.00\pm0.88$ \\
\addlinespace
SICK & Sentence-NLI Transfer & $50.81\pm5.23$ & $46.53\pm2.85$ & $59.33\pm21.79$ & $16.11\pm9.14$ & $77.00\pm4.04$ \\
 & Learned Linker & $\mathbf{56.00\pm0.56}$ & $\mathbf{46.95\pm0.22}$ & $92.56\pm1.84$ & $4.78\pm1.26$ & $70.67\pm1.20$ \\
\bottomrule
\end{tabular}
\caption{Raw relation classification on 3,600 atom pairs: 900 per split, 300 per class (\%). Values are three-seed mean $\pm$ sample standard deviation. Labels generated by a large language model (LLM) were reviewed by an author.}
\label{tab:atom-relation-probes}
\end{table*}

\begin{table*}[t]
\centering
\small
\setlength{\tabcolsep}{7pt}
\begin{tabular}{lrrrr}
\toprule
Data & Representation & Multi-source & Local action & Other/uncertain \\
 & issue & requirement & disagreement & \\
\midrule
SNLI & 269 (53.8) & 59 (11.8) & 287 (57.4) & 90 (18.0) \\
MNLI-m & 434 (86.8) & 52 (10.4) & 328 (65.6) & 30 (6.0) \\
MNLI-mm & 443 (88.6) & 55 (11.0) & 348 (69.6) & 55 (11.0) \\
SICK & 178 (35.6) & 55 (11.0) & 300 (60.0) & 88 (17.6) \\
\midrule
Pooled & 1,324 (66.2) & 221 (11.1) & 1,263 (63.2) & 263 (13.2) \\
\bottomrule
\end{tabular}
\caption{Issues in 2,000 Linker errors (seed 2026; 500 per split). Entries are sentence-pair counts (\%); categories overlap. Pooled percentages weight the four splits equally. LLM-generated annotations were reviewed by an author.}
\label{tab:error-issues}
\end{table*}

We evaluate relation classification on 3,600 ordered pairs of atom
verbalizations, with 300 pairs per class in each split
(Table~\ref{tab:atom-relation-probes}). Selection caps the contribution of each
semantic family, source example, and atom text
(Appendix~\ref{app:atom-relation-construction}). The probe uses the selected
checkpoints without further training; its annotations are not used for
checkpoint or threshold selection. Each model predicts the relation with the
highest Ent/Con/Neu logit for each unchanged atom pair, without source
selection, a no-link action, or SAT-based inference. Some labels assume shared
referents or aligned roles; these assumptions are not supplied to the models.

Learned Linker improves mean probe Accuracy over Transfer by 0.37--9.74
points across splits. Its Contradiction recall is higher on SNLI but lower
on both MNLI splits and SICK. Macro-F1 is higher on SICK
(46.95\% versus 46.53\%). Per-seed Accuracy and Macro-F1 are reported in
Appendix~\ref{app:atom-relation-probes}.

On SNLI and both MNLI splits, Learned Linker's Entailment recall is
96.67--98.67\%, whereas Neutral recall is 29.33--45.33\%.
Under Equation~\ref{eq:choice-scores}, an Entailment or Contradiction logit above
the Neutral logit gives a positive link score, exceeding the no-link score
of zero. For a Neutral pair, this favors linking when the pair is a candidate,
although the selected source depends on all candidate scores.
On SICK, mean Contradiction recall is 4.78\% in the pair probe and 70.79\%
at the sentence level. Formula-level negation can yield sentence-level
contradiction through exact matches or entailment links. Such cases do not
require the linker to predict atom-level contradiction.

\subsection{Error Analysis}
\label{sec:error-analysis}

We inspect 2,000 Learned Linker errors from seed 2026: 500 per split,
stratified by reference label and unmatched-target count. The sample includes
only errors with a valid prediction and at least one target offering multiple
legal actions. Target annotations were produced with reference labels and
model decisions hidden, then compared with the saved choices.

Table~\ref{tab:error-issues} reports four overlapping sentence-level categories.
\emph{Representation issues} indicate that meaning is omitted or distorted
in atoms or formulae. \emph{Multi-source requirements} arise when a target
needs joint support from multiple premise atoms. \emph{Local action
disagreements} occur when a single-source-expressible target's selected action
falls outside its annotated acceptable set. \emph{Other/uncertain} covers ambiguous cases and errors
outside these categories.
Appendix~\ref{app:error-audit} specifies eligibility and counting rules.

Representation issues and local action disagreements are the most frequent
observations (66.2\% and 63.2\%); 40.3\% of sampled errors exhibit both.
Representation issues occur in 86.8\% of the MNLI-m sample and 88.6\% of the
MNLI-mm sample. The audit does not isolate their effects on the prediction.
Appendix~\ref{app:error-cases} gives eight error cases.

\FloatBarrier
\section{Conclusion}
\label{sec:conclusion}

Learned Atom Linker + SAT learns proposition-level links from sentence labels
without atom-pair training annotations, while exposing the atom links and
logical clauses used for each prediction.
With the translator and SAT reasoner fixed, it improves sentence-level accuracy
over Sentence-NLI Transfer on all four evaluation splits.
It also achieves higher accuracy than FH-SAT, with a smaller gain on SICK,
while remaining below Direct Sentence.

\section*{Limitations}

The linker cannot revise the fixed AMR-based representations, including their
errors and omissions. Each unmatched target selects at most one source atom,
so its link action cannot directly encode joint support from multiple premise
atoms. Learned link actions cannot express reverse entailment, which some
inferences under negation require.
Sentence labels can be consistent with multiple link structures, so
correct sentence predictions do not guarantee semantically correct links.
If the Hamming-1 training pool contains no gold-label world, the example
provides no training signal in that epoch
(Appendix~\ref{app:target-complexity}).

The experiments use separate English linkers for SNLI, MNLI, and SICK;
cross-dataset and multilingual transfer are untested.

The balanced probe and the stratified error sample from
one seed do not estimate overall candidate or error frequencies.

\bibliography{references}

\clearpage
\appendix
\section*{Appendix}
\section{Data and Training Configuration}
\label{app:data-selection}

\subsection{Training and Development Data}
For SNLI and MNLI, we randomly sample 30,000 examples per official training
split (10,000 per label; seed 1129).
Eligible examples have valid propositional representations and at least one
unmatched target with multiple legal actions. We retain one copy of same-label
duplicate sentence pairs and exclude pairs with conflicting labels. Training excludes
development IDs, sentence pairs, and individual sentences, as well as
evaluation IDs and complete sentence pairs.

For SICK, Direct Sentence and Learned Linker use the same 3,722 eligible
training examples without class balancing.
Evaluation uses the 4,906-example public test split from the
Hugging Face \texttt{sick/default} release.

All systems use the same development sets: 3,000 examples from SNLI validation,
3,000 held out from MNLI training, and all 495 SICK validation examples.
The SNLI and MNLI development sets contain 1,000 examples per label.

\subsection{Hyperparameter Search and Selected Configuration}
\label{app:hyperparameters}
\begin{table*}[!t]
\centering
\small
\setlength{\tabcolsep}{3pt}
\begin{tabular}{lcccccc}
\toprule
& \multicolumn{3}{c}{Direct Sentence} & \multicolumn{3}{c}{Learned Linker} \\
\cmidrule(lr){2-4}\cmidrule(lr){5-7}
Setting & SNLI & MNLI & SICK & SNLI & MNLI & SICK \\
\midrule
Encoder LR candidates & \multicolumn{6}{c}{$\{5\times10^{-6},2\times10^{-5}\}$} \\
Effective batch candidates & \multicolumn{6}{c}{$\{64,128\}$} \\
\addlinespace
Encoder LR & \multicolumn{6}{c}{$2\times10^{-5}$} \\
Head LR ($10\times$ encoder) & \multicolumn{6}{c}{$2\times10^{-4}$} \\
Effective batch size & 64 & 64 & 128 & 128 & 64 & 128 \\
Input cap (tokens) & \multicolumn{3}{c}{256} & \multicolumn{3}{c}{64} \\
\midrule
\multicolumn{7}{l}{\emph{Fixed settings}} \\
Encoder & \multicolumn{6}{c}{MiniLMv1-L12-H384 (33M)} \\
Classifier head & \multicolumn{6}{c}{Final [CLS] $\rightarrow$ 384-unit tanh layer $\rightarrow$ 3-way linear layer} \\
Classifier dropout & \multicolumn{6}{c}{0.1 before each linear layer} \\
Warmup fraction & \multicolumn{6}{c}{0.10} \\
Weight decay & \multicolumn{6}{c}{0.01} \\
Gradient clipping ($\ell_2$) & \multicolumn{6}{c}{1.0} \\
Early-stop patience & 2 & 2 & 5 & 2 & 2 & 5 \\
Maximum epochs & 10 & 10 & 20 & 10 & 10 & 20 \\
Micro-batch size & \multicolumn{3}{c}{32} & \multicolumn{3}{c}{8} \\
Micro-batch pair budget & \multicolumn{3}{c}{---} & \multicolumn{3}{c}{96} \\
\bottomrule
\end{tabular}
\caption{Selected training settings and search ranges. All six searches use
seed 2026. Effective batch size counts examples per optimizer update.}
\label{tab:optimization}
\label{tab:hyperparameter-grid}
\label{tab:direct-optimization}
\end{table*}

Direct Sentence and Learned Linker use
\texttt{microsoft/MiniLM-L12-H384-uncased}.
Table~\ref{tab:optimization} reports the selected settings and search ranges,
and Table~\ref{tab:selected-dev} reports the selected epochs and development scores.
For each system and dataset, we search the four combinations of two encoder
learning rates and two effective batch sizes (seed 2026). We select
configurations and checkpoints by development Macro-F1, breaking ties by
Accuracy, without using evaluation splits.
Main evaluation reuses the selected seed-2026 checkpoints; seeds
2027 and 2028 use the same configurations.

Both systems use AdamW with linear warmup and decay.
Linker micro-batches contain at most eight examples and 96 candidate atom
pairs; an example exceeding the pair limit is processed alone.
Linker training uses gradient checkpointing.

Direct Sentence and Transfer use a 256-token input cap; Learned Linker uses
64. Caps include both texts and special tokens. No candidate atom pair is
truncated on any full evaluation split; maximum lengths are 21, 35, 34, and
19 tokens on SNLI, MNLI-m, MNLI-mm, and SICK, respectively.

\begin{table*}[!t]
\centering
\footnotesize
\setlength{\tabcolsep}{3pt}
\begin{tabular}{llrrrrrrrrr}
\toprule
Data & System & \multicolumn{3}{c}{2026} & \multicolumn{3}{c}{2027} & \multicolumn{3}{c}{2028} \\
\cmidrule(lr){3-5}\cmidrule(lr){6-8}\cmidrule(lr){9-11}
 & & Epoch & Acc. & M-F1 & Epoch & Acc. & M-F1 & Epoch & Acc. & M-F1 \\
\midrule
SNLI & Direct Sentence & 7 & 85.87 & 85.83 & 6 & 86.80 & 86.83 & 7 & 86.70 & 86.68 \\
 & Learned Linker & 8 & 80.07 & 80.06 & 8 & 80.03 & 80.06 & 6 & 79.83 & 79.88 \\
\addlinespace
MNLI & Direct Sentence & 4 & 78.97 & 79.00 & 5 & 79.07 & 79.00 & 4 & 79.50 & 79.48 \\
 & Learned Linker & 8 & 60.57 & 60.79 & 10 & 60.80 & 61.08 & 8 & 60.00 & 60.32 \\
\addlinespace
SICK & Direct Sentence & 11 & 87.68 & 87.13 & 11 & 87.27 & 86.92 & 17 & 87.47 & 87.32 \\
 & Learned Linker & 15 & 83.43 & 81.21 & 17 & 82.02 & 80.01 & 17 & 81.21 & 79.52 \\
\bottomrule
\end{tabular}
\caption{Selected epochs and development Accuracy and Macro-F1 (\%) for the 18 neural runs in the main experiment. Transfer reuses the corresponding Direct Sentence checkpoint; MNLI-m and MNLI-mm share one MNLI checkpoint per system and seed.}
\label{tab:selected-dev}
\end{table*}

\paragraph{FH-SAT.}
We freeze the embedding model \texttt{BAAI/bge-small-en-v1.5}
\citep{xiao-etal-2024-cpack} and the
NLI model \texttt{MoritzLaurer/\allowbreak mDeBERTa-\allowbreak v3-\allowbreak base-\allowbreak xnli-\allowbreak multilingual-\allowbreak nli-\allowbreak 2mil7}
\citep{he-etal-2023-debertav3,laurer-etal-2024-less}.
We search similarity thresholds $\tau_m\in\{0,0.05,\ldots,1\}$ and NLI
confidence thresholds $\tau_c\in\{80,90,100\}$ (percent), selecting by
development Macro-F1, then Accuracy, then lower $\tau_m$ and $\tau_c$.
Table~\ref{tab:fh-settings} reports the development sizes and selected values.
MNLI's matched and mismatched evaluations share the selected thresholds.

\begin{table}[!htbp]
\centering
\small
\begin{tabular}{lrrr}
\toprule
Dataset & Dev examples & $\tau_m$ & $\tau_c$ (\%) \\
\midrule
SNLI & 3,000 & 0.75 & 80 \\
MNLI & 3,000 & 0.65 & 90 \\
SICK & 495 & 0.85 & 100 \\
\bottomrule
\end{tabular}
\caption{FH-SAT development sizes and selected thresholds.}
\label{tab:fh-settings}
\end{table}

\section{Atom-Relation Probe}
\label{app:atom-relation-construction}
\subsection{Construction and Annotation}

The probe contains 300 distinct ordered atom pairs per relation in each split.
Table~\ref{tab:pair-composition} lists the number of pairs in each semantic family.

Within each combination of split and relation, each original example occurs in at most two pairs,
and each atom text occurs at most three times across both endpoints.
Ordered atom-text pairs are unique across splits and relations. If either
direction is labeled Contradiction, only one orientation is retained.
Neutral pairs are premise-to-hypothesis candidates from the same example.
Entailment and Contradiction may use atoms from different
examples under stated alignment conditions.

OpenAI's GPT-6 labels retrieved atom pairs under specified alignment
conditions. Annotation items omit parent sentences, benchmark gold labels,
model predictions, and the relation categories used to retrieve candidates.
An author reviewed all 3,600 pair annotations.
Appendix~\ref{app:pair-prompt} gives the initial annotation prompt.

\subsection{Evaluation and Per-Seed Results}
\label{app:atom-relation-probes}
All checkpoint evaluations use the same pairs and labels.
Unchanged atom texts are tokenized with caps of 64 for Learned Linker and 256
for Transfer; no pair is truncated. Some labels assume a shared referent,
shared event, or aligned numerical slot; these conditions are not model inputs.
Table~\ref{tab:per-seed-atom-relations} reports descriptive per-seed results
without pair-level significance tests.

\clearpage
\begin{table}[H]
\centering
\footnotesize
\setlength{\tabcolsep}{2pt}
\begin{tabular}{@{}lrrrr@{}}
\toprule
Semantic family & SNLI & MNLI-m & MNLI-mm & SICK \\
\midrule
\multicolumn{5}{l}{\textbf{Entailment}} \\
Argument projection & 60 & 68 & 63 & 60 \\
Event entailment & 60 & 68 & 67 & 60 \\
Modifier generalization & 60 & 40 & 34 & 60 \\
Nominal entailment & 60 & 59 & 65 & 60 \\
Paraphrase & 60 & 65 & 71 & 60 \\
Total & 300 & 300 & 300 & 300 \\
\midrule
\multicolumn{5}{l}{\textbf{Contradiction}} \\
Attitude preference & 1 & 25 & 32 & 0 \\
Clock/date-time & 0 & 15 & 15 & 0 \\
Color & 58 & 13 & 4 & 57 \\
Event opposition & 7 & 34 & 41 & 19 \\
Exact quantity & 60 & 48 & 58 & 60 \\
Permission decision & 0 & 2 & 1 & 0 \\
Posture & 60 & 31 & 14 & 75 \\
Property state & 50 & 73 & 71 & 38 \\
Relational opposition & 0 & 4 & 5 & 0 \\
Spatial direction & 59 & 25 & 16 & 31 \\
Verbalized negation & 5 & 30 & 43 & 20 \\
Total & 300 & 300 & 300 & 300 \\
\midrule
\multicolumn{5}{l}{\textbf{Neutral}} \\
Additional content & 75 & 65 & 69 & 75 \\
Event mismatch & 64 & 45 & 44 & 63 \\
Modifier mismatch & 9 & 8 & 10 & 9 \\
Compatible quantity & 5 & 15 & 19 & 7 \\
Reverse entailment & 15 & 2 & 1 & 8 \\
Role entity mismatch & 61 & 75 & 75 & 75 \\
Scope mismatch & 9 & 25 & 22 & 1 \\
Unrelated & 62 & 65 & 60 & 62 \\
Total & 300 & 300 & 300 & 300 \\
\bottomrule
\end{tabular}
\caption{Atom-pair counts by relation, semantic family, and split. Within each split and relation, the family cap is 75, except for color and exact-quantity conflicts (60 each) and clock/date-time conflicts (15).}
\label{tab:pair-composition}
\end{table}

\begin{table}[H]
\centering
\footnotesize
\setlength{\tabcolsep}{2.2pt}
\begin{tabular}{lllrrr}
\toprule
Data & System & Metric & 2026 & 2027 & 2028 \\
\midrule
SNLI & Transfer & Acc. & 63.00 & 65.78 & 64.78 \\
 &  & M-F1 & 57.15 & 60.82 & 58.67 \\
 & Linker & Acc. & 72.33 & 76.22 & 74.22 \\
 &  & M-F1 & 70.95 & 74.73 & 73.08 \\
\addlinespace
MNLI-m & Transfer & Acc. & 66.22 & 66.67 & 66.44 \\
 &  & M-F1 & 62.99 & 63.08 & 61.97 \\
 & Linker & Acc. & 66.78 & 69.11 & 67.44 \\
 &  & M-F1 & 63.25 & 65.91 & 64.61 \\
\addlinespace
MNLI-mm & Transfer & Acc. & 69.00 & 68.78 & 67.56 \\
 &  & M-F1 & 65.84 & 65.03 & 62.59 \\
 & Linker & Acc. & 68.67 & 70.00 & 67.78 \\
 &  & M-F1 & 65.16 & 66.61 & 64.74 \\
\addlinespace
SICK & Transfer & Acc. & 44.78 & 53.89 & 53.78 \\
 &  & M-F1 & 43.25 & 48.40 & 47.96 \\
 & Linker & Acc. & 56.56 & 56.00 & 55.44 \\
 &  & M-F1 & 46.81 & 47.21 & 46.84 \\
\bottomrule
\end{tabular}
\caption{Per-seed Accuracy and Macro-F1 on the 3,600-pair probe (\%); each split contains 900 pairs.}
\label{tab:per-seed-atom-relations}
\end{table}

\newpage
\subsection{Initial Annotation Prompt}
\label{app:pair-prompt}

\begingroup
\small\ttfamily\raggedright
\setlength{\parindent}{0pt}
\setlength{\smallskipamount}{2pt plus 1pt minus 1pt}
\noindent \# Blind semantic review of emitted atom pairs\par

\smallskip
\noindent Judge the directed relation from source to target using only their actual verbalizations and explicitly provided alignment conditions. Do not inspect parent sentences, AMR, formulae, proposed class/family, dataset, quota progress, old annotations, sentence gold or any model output. Never repair an atom from its parent sentence or assume an unexpressed attribute or argument.\par

\smallskip
\noindent Ent: source entails target. Con: incompatible under the stated conditions. Neu: neither. Reject invalid provenance/construction; ambiguous means a material reading cannot be resolved from the allowed inputs. Awkward but interpretable grammar alone is not rejection. Preserve roles, names, quantities, verbalized negation, modality and direction. Outer formula polarity is not an atom relation. Two different activities/objects are not automatically incompatible. Different colors require a coherent single-color reading; different numbers require an exact reading of the same set/slot. Conditions are not evidence that a label already holds. Check whether the words actually express the specified slot. An event argument may be dropped only if the target's predicate, sense, roles and restrictions survive; do not assume all longer atoms entail shorter ones. WordNet or lexical retrieval, if any, is not an annotation authority.\par

\smallskip
\noindent Return candidate\_id, relation (ent/con/neu/ambiguous/rejected), semantic\_family, confidence (high/medium/low), concrete rationale, and alignment\_concern. Choose the family by the actual semantic reason, not surface length alone:\par

\smallskip
\noindent Ent families:\par

\smallskip
\noindent - argument\_projection: same event loses an argument/adjunct, no other inference.\par

\smallskip
\noindent - modifier\_generalization: removal/generalization of an entity/property modifier.\par

\smallskip
\noindent - nominal\_entailment: noun/category inclusion, preserving the rest of the claim.\par

\smallskip
\noindent - event\_entailment: a more specific action/state entails a broader action/state.\par

\smallskip
\noindent - paraphrase: semantically equivalent lexical, morphological or syntactic form.\par

\smallskip
\noindent Con families:\par

\smallskip
\noindent - color; exact\_quantity; clock\_time; posture; spatial\_direction; property\_state; event\_opposition; attitude\_preference; permission\_decision; relational\_opposition; verbalized\_negation. Use clock\_time for hour/minute/date-slot conflicts; exact\_quantity for other exact numerical and quantifier oppositions. Avoid splitting one family simply because its entity, number, adjective or predicate changes.\par

\smallskip
\noindent Neu families (choose the principal reason):\par

\smallskip
\noindent - unrelated: different content with no substantive shared semantic claim.\par

\smallskip
\noindent - additional\_content: target adds a claim not supplied by source.\par

\smallskip
\noindent - role\_entity\_mismatch: participant identity or role arrangement differs without establishing incompatibility.\par

\smallskip
\noindent - event\_mismatch: events differ without known entailment or incompatibility.\par

\smallskip
\noindent - modifier\_mismatch: attributes differ but may coexist or lack common scope.\par

\smallskip
\noindent - quantity\_compatible: numbers/comparisons need not conflict under actual words.\par

\smallskip
\noindent - reverse\_entailment: target would entail source, but not source to target.\par

\smallskip
\noindent - scope\_mismatch: negation/modality/scope blocks entailment without proving Con.\par

\smallskip
\noindent Use semantic\_family=unresolved for ambiguous/rejected. Keep all reclassifications and rejects. Never change standards or family to fill a quota. These are assistant first-pass judgements, not completed human review.\par
\endgroup

\FloatBarrier

\section{Error Analysis}
\label{app:error-audit}
\subsection{Sampling and Annotation}

We sample 2,000 seed-2026 Learned Linker errors, 500 per split, excluding
Invalid outputs and requiring at least one target with multiple legal actions.
Eligible pools contain 2,121 SNLI, 3,798 MNLI-m, 3,741 MNLI-mm, and 748 SICK
errors. We allocate sample quotas proportionally across reference-label and
unmatched-target-count strata, selecting deterministically within each
stratum. Pooled percentages give equal weight to the four sampled error sets.

OpenAI's GPT-6 annotated the unmatched targets using the original sentences, AMRs,
formulae, and legal actions, with reference labels and model outputs hidden.
An author reviewed all 2,000 examples; the annotations were frozen before
comparison with the model's selected actions.

\emph{Target annotations.} We record expressibility, rationale, confidence, and, where
applicable, acceptable actions and supporting atoms. Expressibility is
classified as single-source-expressible, multi-source-required,
representation-inadequate, or ambiguous.
Single-source-expressible includes targets whose only acceptable action is
no-link. Annotations distinguish positive atom meanings from outer negation
and do not treat disjunctive branches as jointly true.

\emph{Sentence-pair flags.} We summarize the target annotations and
example-level notes with four binary flags per sentence pair. The flags may overlap.

\emph{Representation issues}: an inadequate target representation or a
defect recorded in the example-level notes, including omitted or
exact-matched material.

\emph{Multi-source requirements}: a target is annotated as needing joint
support from multiple sources.

\emph{Local action disagreements}: a single-source-expressible target's
selected action falls outside its annotated acceptable set; no-link is
included in this comparison.

\emph{Other/uncertain}: an ambiguous target, uncertainty in the example-level
notes, or the absence of all three preceding flags.
Appendix~\ref{app:error-prompt} gives the initial annotation prompt;
review and aggregation procedures accompany the code.

\subsection{Illustrative Error Cases}
\label{app:error-cases}
The cases use seed-2026 predictions and retain the released identifiers, atom indices, and verbalizations. Within each case, $s_i$ and $t_i$ denote premise and hypothesis atoms. Link clauses follow Definition~\ref{def:world-theory}; only X000435 contains exact matches.

\paragraph{Location changed into activity (X000015).}
SNLI; reference/prediction: Con/Ent.
\par\nopagebreak\noindent\emph{Premise:} ``A police person is on a motorcycle on the side of a street.''
\par\noindent\emph{Hypothesis:} ``A person is on a bike.''
\par\noindent\emph{Premise atoms:}\par\nopagebreak\noindent
\begin{tabular}{@{}l@{\quad}p{0.87\linewidth}@{}}
$s_{1}$ & side is part of street \\
$s_{2}$ & person have org role police \\
$s_{3}$ & person be located at at motorcycle \\
$s_{4}$ & person be located at at side \\
\end{tabular}
\par\noindent\emph{Hypothesis atoms:}\par\nopagebreak\noindent
\begin{tabular}{@{}l@{\quad}p{0.87\linewidth}@{}}
$t_{1}$ & person bike \\
\end{tabular}
\[
\begin{aligned}
\Phi_P &= s_{3} \land s_{4} \land s_{2} \land s_{1},\\
\Phi_H &= t_{1}.
\end{aligned}
\]
\par\noindent\emph{Selected actions:} Ent from $s_{3}$ to $t_{1}$ ($s_{3}\to t_{1}$).
\par\noindent\emph{Why the reasoner predicts Ent:} The premise and selected $s_3\to t_1$ entail $\Phi_H=t_1$.
\par\noindent\emph{Error analysis:} The hypothesis describes a location, but its AMR uses the activity predicate \emph{bike-01}, so $t_1$ describes biking. The annotation marks $t_1$ as representation-inadequate and excludes it from local action-disagreement scoring. This change precedes linking; it does not establish that the defect alone explains the benchmark label, whose interpretation also depends on ``bike'' versus ``motorcycle.''

\paragraph{Different percentages with identical verbalizations (X000435).}
MNLI-m; reference/prediction: Con/Ent.
\par\nopagebreak\noindent\emph{Premise:} ``yeah uh yeah absolutely and the credit union has nine percent interest so yeah so that's''
\par\noindent\emph{Hypothesis:} ``The credit union has one hundred percent interest so yeah.''
\par\noindent\emph{Premise atoms:}\par\nopagebreak\noindent
\begin{tabular}{@{}l@{\quad}p{0.87\linewidth}@{}}
$s_{1}$ & percentage entity interest \\
$s_{2}$ & credit union \\
$s_{3}$ & yeah has expressive mode \\
$s_{4}$ & yeah absolute have \\
$s_{5}$ & union have interest \\
$s_{6}$ & absolute cause that \\
$s_{7}$ & have cause that \\
\end{tabular}
\par\noindent\emph{Hypothesis atoms:}\par\nopagebreak\noindent
\begin{tabular}{@{}l@{\quad}p{0.87\linewidth}@{}}
$t_{1}$ & percentage entity interest \\
$t_{2}$ & credit union \\
$t_{3}$ & union have interest \\
$t_{4}$ & have cause yeah \\
\end{tabular}
\[
\begin{aligned}
\Phi_P &= s_{4} \land s_{6} \land s_{5} \land s_{7} \land s_{2} \land s_{1} \land s_{3},\\
\Phi_H &= t_{3} \land t_{2} \land t_{1} \land t_{4}.
\end{aligned}
\]
\par\noindent\emph{Exact matches:} $s_{1}\leftrightarrow t_{1}$, $s_{2}\leftrightarrow t_{2}$, $s_{5}\leftrightarrow t_{3}$.
\par\noindent\emph{Selected actions:} Ent from $s_{7}$ to $t_{4}$ ($s_{7}\to t_{4}$).
\par\noindent\emph{Why the reasoner predicts Ent:} With all source atoms asserted, exact matches establish $t_1$, $t_2$, and $t_3$; the selected $s_7\to t_4$ establishes $t_4$. Thus all conjuncts of $\Phi_H$ hold.
\par\noindent\emph{Error analysis:} The raw AMRs and structured atom metadata retain 9 and 100, but $s_1$ and $t_1$ both verbalize as \emph{percentage entity interest}. Exact matching equates them, leaving the numerical contrast out of the SAT theory. For the unmatched $t_4$, the annotation accepts no-link because ``that'' in $s_7$ does not establish ``yeah.'' The audit counts the saved Ent action as a local action disagreement.

\paragraph{Event content and instrument (X000019).}
SICK; reference/prediction: Ent/Neu.
\par\nopagebreak\noindent\emph{Premise:} ``A man is making music with a flute''
\par\noindent\emph{Hypothesis:} ``A man is playing flute''
\par\noindent\emph{Premise atoms:}\par\nopagebreak\noindent
\begin{tabular}{@{}l@{\quad}p{0.87\linewidth}@{}}
$s_{1}$ & man make music \\
$s_{2}$ & man make using flute \\
\end{tabular}
\par\noindent\emph{Hypothesis atoms:}\par\nopagebreak\noindent
\begin{tabular}{@{}l@{\quad}p{0.87\linewidth}@{}}
$t_{1}$ & man play using flute \\
\end{tabular}
\[
\begin{aligned}
\Phi_P &= s_{1} \land s_{2},\\
\Phi_H &= t_{1}.
\end{aligned}
\]
\par\noindent\emph{Selected actions:} no-link for $t_{1}$.
\par\noindent\emph{Why the reasoner predicts Neu:} With no exact match and a no-link choice, $\Phi_{\hat w}=s_1\land s_2$ permits both $t_1$ and $\neg t_1$.
\par\noindent\emph{Error analysis:} In the shared event, $s_1$ identifies the product as music and $s_2$ the instrument as a flute. The annotation marks $t_1$ as multi-source-required: the support $(s_1\land s_2)\to t_1$ is outside the single-source action space.

\paragraph{A jump and its path landmark (F000952).}
SICK; reference/prediction: Ent/Neu.
\par\nopagebreak\noindent\emph{Premise:} ``A deer is jumping over a fence''
\par\noindent\emph{Hypothesis:} ``A deer is jumping a fence''
\par\noindent\emph{Premise atoms:}\par\nopagebreak\noindent
\begin{tabular}{@{}l@{\quad}p{0.87\linewidth}@{}}
$s_{1}$ & over fence \\
$s_{2}$ & deer jump along over \\
\end{tabular}
\par\noindent\emph{Hypothesis atoms:}\par\nopagebreak\noindent
\begin{tabular}{@{}l@{\quad}p{0.87\linewidth}@{}}
$t_{1}$ & deer jump fence \\
\end{tabular}
\[
\begin{aligned}
\Phi_P &= s_{2} \land s_{1},\\
\Phi_H &= t_{1}.
\end{aligned}
\]
\par\noindent\emph{Selected actions:} no-link for $t_{1}$.
\par\noindent\emph{Why the reasoner predicts Neu:} With no exact match and a no-link choice, $\Phi_{\hat w}=s_2\land s_1$ permits both $t_1$ and $\neg t_1$.
\par\noindent\emph{Error analysis:} The atoms share an \emph{over} node in the raw AMR: $s_1$ supplies the fence landmark and $s_2$ the deer's jump. The annotation marks $t_1$ as multi-source-required: the support $(s_1\land s_2)\to t_1$ is outside the single-source action space.

\paragraph{A link where no-link is acceptable (X001174).}
SICK; reference/prediction: Neu/Ent.
\par\nopagebreak\noindent\emph{Premise:} ``A man is crying''
\par\noindent\emph{Hypothesis:} ``A man is screaming''
\par\noindent\emph{Premise atoms:}\par\nopagebreak\noindent
\begin{tabular}{@{}l@{\quad}p{0.87\linewidth}@{}}
$s_{1}$ & man cry \\
\end{tabular}
\par\noindent\emph{Hypothesis atoms:}\par\nopagebreak\noindent
\begin{tabular}{@{}l@{\quad}p{0.87\linewidth}@{}}
$t_{1}$ & man scream \\
\end{tabular}
\[
\begin{aligned}
\Phi_P &= s_{1},\\
\Phi_H &= t_{1}.
\end{aligned}
\]
\par\noindent\emph{Selected actions:} Ent from $s_{1}$ to $t_{1}$ ($s_{1}\to t_{1}$).
\par\noindent\emph{Why the reasoner predicts Ent:} The selected world $\Phi_{\hat w}=s_1\land(\neg s_1\lor t_1)$ is consistent and entails $t_1$.
\par\noindent\emph{Error analysis:} Crying need not involve screaming, and the activities can coexist, so neither an entailment nor a contradiction link is justified. The annotation accepts only no-link; the selected Ent action adds an unsupported implication.

\paragraph{Entailment instead of an acceptable contradiction link (X001873).}
SICK; reference/prediction: Con/Ent.
\par\nopagebreak\noindent\emph{Premise:} ``A man is speaking''
\par\noindent\emph{Hypothesis:} ``A man is silent''
\par\noindent\emph{Premise atoms:}\par\nopagebreak\noindent
\begin{tabular}{@{}l@{\quad}p{0.87\linewidth}@{}}
$s_{1}$ & man speak \\
\end{tabular}
\par\noindent\emph{Hypothesis atoms:}\par\nopagebreak\noindent
\begin{tabular}{@{}l@{\quad}p{0.87\linewidth}@{}}
$t_{1}$ & man is silent \\
\end{tabular}
\[
\begin{aligned}
\Phi_P &= s_{1},\\
\Phi_H &= t_{1}.
\end{aligned}
\]
\par\noindent\emph{Selected actions:} Ent from $s_{1}$ to $t_{1}$ ($s_{1}\to t_{1}$).
\par\noindent\emph{Why the reasoner predicts Ent:} The consistent world asserts $s_1$ and includes $s_1\to t_1$, so it entails $\Phi_H=t_1$.
\par\noindent\emph{Error analysis:} For the same participant and moment, speaking is incompatible with being silent. The annotation accepts Con from $s_1$ to $t_1$, represented by $\neg(s_1\land t_1)$.

\paragraph{Missing AM/PM interpretation (X001162).}
MNLI-mm; reference/prediction: Con/Neu.
\par\nopagebreak\noindent\emph{Premise:} ``The time was 9:38.''
\par\noindent\emph{Hypothesis:} ``The time was after noon.''
\par\noindent\emph{Premise atoms:}\par\nopagebreak\noindent
\begin{tabular}{@{}l@{\quad}p{0.87\linewidth}@{}}
$s_{1}$ & date entity occurs at 9:38 \\
\end{tabular}
\par\noindent\emph{Hypothesis atoms:}\par\nopagebreak\noindent
\begin{tabular}{@{}l@{\quad}p{0.87\linewidth}@{}}
$t_{1}$ & be temporally at occurs during after \\
$t_{2}$ & after noon \\
\end{tabular}
\[
\begin{aligned}
\Phi_P &= s_{1},\\
\Phi_H &= t_{1} \land t_{2}.
\end{aligned}
\]
\par\noindent\emph{Selected actions:} Ent from $s_{1}$ to $t_{1}$ ($s_{1}\to t_{1}$); no-link for $t_{2}$.
\par\noindent\emph{Why the reasoner predicts Neu:} The premise and selected $s_1\to t_1$ establish $t_1$, while $t_2$ remains unconstrained. Both $\Phi_H=t_1\land t_2$ and its negation are consistent with $\Phi_{\hat w}$.
\par\noindent\emph{Error analysis:} Without an AM/PM convention, 9:38 can fall before or after noon. Both unmatched targets are annotated as ambiguous, with no acceptable action sets assigned. The mismatch with the benchmark's Con label therefore does not establish that either selected action is a linking error.

\paragraph{A missing implication in the opposite direction (X001167).}
SICK; reference/prediction: Con/Neu.
\par\nopagebreak\noindent\emph{Premise:} ``The man is not exercising''
\par\noindent\emph{Hypothesis:} ``A man is doing pull-ups''
\par\noindent\emph{Premise atoms:}\par\nopagebreak\noindent
\begin{tabular}{@{}l@{\quad}p{0.87\linewidth}@{}}
$s_{1}$ & man exercise \\
\end{tabular}
\par\noindent\emph{Hypothesis atoms:}\par\nopagebreak\noindent
\begin{tabular}{@{}l@{\quad}p{0.87\linewidth}@{}}
$t_{1}$ & man pull up \\
$t_{2}$ & man do pull up \\
\end{tabular}
\[
\begin{aligned}
\Phi_P &= \neg (s_{1}),\\
\Phi_H &= t_{2} \land t_{1}.
\end{aligned}
\]
\par\noindent\emph{Selected actions:} no-link for $t_{1}$; no-link for $t_{2}$.
\par\noindent\emph{Why the reasoner predicts Neu:} With both targets unlinked, $\Phi_{\hat w}=\neg s_1$ constrains neither $t_1$ nor $t_2$. It entails neither $\Phi_H$ nor $\neg\Phi_H$.
\par\noindent\emph{Error analysis:} The negation is in $\Phi_P$, while $s_1$ denotes exercising. Exercising entails neither doing pull-ups nor not doing pull-ups, so both saved no-link actions agree with the annotations. The reverse implication $t_2\to s_1$, for example, would contradict $\Phi_H$ when combined with $\neg s_1$. Allowed links $s_1\to t_j$ and $s_1\to\neg t_j$ impose no target constraint when $s_1$ is false. This limitation is missed by the local action-disagreement criterion, leaving the case in Other/uncertain despite its locally acceptable choices.

\subsection{Initial Annotation Prompt}
\label{app:error-prompt}

The error prompt below is abridged to the annotation instructions.

\begingroup
\small\ttfamily\raggedright
\setlength{\parindent}{0pt}
\setlength{\smallskipamount}{2pt plus 1pt minus 1pt}
\noindent \# Error-conditioned representation/action audit\par

\smallskip
\noindent Annotate every unmatched target in each supplied example. See its premise and hypothesis, raw AMRs, complete fixed formulae, emitted atoms and legal actions. Do not use the benchmark gold label, predicted label, model-selected actions, previous annotations or model scores. The cohort consists of errors; this fact is known, but it must not dictate the semantic annotation or the needed answer.\par

\smallskip
\noindent Judge E(source) when the positive source proposition alone entails the positive target, C(source) when it entails target negation, and no-link when no single legal source provides either relation. List every acceptable action, not one preferred witness. Original context may resolve reference, event identity and scope but cannot silently provide omitted content to a single source atom. Outer formula NOT does not reverse local E/C; verbalized negation remains.\par

\smallskip
\noindent Assess coverage separately from mere wording fluency:\par

\smallskip
\noindent - one\_source\_expressible: at least one justified legal single-source action, including ordinary no-link;\par

\smallskip
\noindent - multi\_source\_required: no source works alone, but named source atoms jointly support/refute the target under a shared, licensed interpretation. Name the sources and explain the composition; do not combine mutually alternative OR branches, incompatible scopes or nonasserted content as actual facts;\par

\smallskip
\noindent - representation\_inadequate: identify a concrete omission, role/sense change, scope or formula defect that materially changes the intended proposition. Cite the displayed sentence and AMR/formula/atom evidence. Awkward language alone is not enough. Preserve additional overlapping issue flags;\par

\smallskip
\noindent - ambiguous\_unjudgeable: unresolved sentence meaning or reference prevents a defensible judgement. Do not force an answer from knowing this is an error.\par

\smallskip
\noindent For every target return group\_id, expressibility, all acceptable\_choice\_indices (nonempty only for one\_source\_expressible), concise rationale, confidence, representation\_note, multi\_source\_support\_ids and overlapping\_issue\_flags. Do not invent new atoms/actions or change the formula. Report omissions with no corresponding target in an additional example-level note. An observed defect does not by itself prove that it caused the final prediction error.\par

\smallskip
\noindent The fixed sample is 2,000 seed-2026 valid, non-exact-only errors: 500 per public split. Sampling is proportional to (gold label, unmatched-target count) strata. The sampler accesses gold/predictions; the semantic annotation view does not. These are error-conditioned diagnostics, not whole-test accuracy, unique causal explanations or a representation upper bound. Human review is separate from this assistant first pass.\par
\endgroup

\section{Comparison Statistics and Benchmark Context}
\label{app:paired}
\subsection{Paired Main-Task Comparisons}
\begin{table*}[!t]
\centering
\footnotesize
\setlength{\tabcolsep}{4pt}
\begin{tabular}{llrrrrrr}
\toprule
Data & System & \multicolumn{2}{c}{2026} & \multicolumn{2}{c}{2027} & \multicolumn{2}{c}{2028} \\
\cmidrule(lr){3-4}\cmidrule(lr){5-6}\cmidrule(lr){7-8}
 & & Acc. & M-F1 & Acc. & M-F1 & Acc. & M-F1 \\
\midrule
SNLI & Direct Sentence & 84.82 & 84.68 & 85.49 & 85.46 & 85.38 & 85.28 \\
 & Transfer & 48.81 & 42.12 & 51.24 & 45.52 & 53.33 & 47.59 \\
 & Learned Linker & 78.38 & 78.29 & 78.41 & 78.36 & 77.87 & 77.85 \\
\addlinespace
MNLI-m & Direct Sentence & 78.79 & 78.76 & 79.62 & 79.51 & 78.91 & 78.82 \\
 & Transfer & 39.08 & 37.30 & 40.57 & 39.09 & 38.57 & 36.46 \\
 & Learned Linker & 60.72 & 60.90 & 61.88 & 62.11 & 61.58 & 61.82 \\
\addlinespace
MNLI-mm & Direct Sentence & 80.23 & 80.19 & 80.52 & 80.38 & 80.07 & 79.97 \\
 & Transfer & 38.63 & 36.64 & 40.71 & 39.19 & 37.93 & 35.55 \\
 & Learned Linker & 61.59 & 61.71 & 62.37 & 62.54 & 62.04 & 62.27 \\
\addlinespace
SICK & Direct Sentence & 87.42 & 86.46 & 87.18 & 86.43 & 86.69 & 86.03 \\
 & Transfer & 61.48 & 56.95 & 78.54 & 74.37 & 78.29 & 74.60 \\
 & Learned Linker & 84.45 & 82.51 & 84.02 & 82.19 & 84.04 & 82.18 \\
\bottomrule
\end{tabular}
\caption{Per-seed Accuracy and Macro-F1 on the full public evaluation splits (\%). Transfer denotes Sentence-NLI Transfer. Deterministic FH-SAT results appear in Table~\ref{tab:main-results}.}
\label{tab:main-per-seed}
\end{table*}

\begin{table*}[!t]
\centering
\footnotesize
\setlength{\tabcolsep}{4pt}
\begin{tabular}{llrrrr}
\toprule
Comp. & Data & Acc. $\Delta$ & 95\% CI & M-F1 $\Delta$ & 95\% CI \\
\midrule
L--T & SNLI & $27.09\pm2.52$ & $[24.53,29.61]$ & $33.09\pm2.96$ & $[30.28,36.07]$ \\
L--T & MNLI-m & $21.99\pm0.90$ & $[20.62,23.42]$ & $23.99\pm1.22$ & $[22.47,25.64]$ \\
L--T & MNLI-mm & $22.91\pm1.23$ & $[21.32,24.54]$ & $25.05\pm1.69$ & $[23.15,26.96]$ \\
L--T & SICK & $11.40\pm10.02$ & $[4.97,22.73]$ & $13.65\pm10.31$ & $[6.97,25.29]$ \\
\addlinespace
L--FH & SNLI & $26.32\pm0.30$ & $[25.15,27.49]$ & $29.51\pm0.28$ & $[28.36,30.69]$ \\
L--FH & MNLI-m & $8.14\pm0.60$ & $[6.95,9.33]$ & $6.75\pm0.63$ & $[5.55,7.94]$ \\
L--FH & MNLI-mm & $8.51\pm0.39$ & $[7.45,9.59]$ & $6.96\pm0.42$ & $[5.87,8.02]$ \\
L--FH & SICK & $2.07\pm0.24$ & $[1.06,3.06]$ & $1.75\pm0.19$ & $[0.64,2.86]$ \\
\addlinespace
L--D & SNLI & $-7.01\pm0.54$ & $[-7.93,-6.12]$ & $-6.97\pm0.53$ & $[-7.89,-6.08]$ \\
L--D & MNLI-m & $-17.71\pm0.37$ & $[-18.79,-16.66]$ & $-17.42\pm0.43$ & $[-18.51,-16.36]$ \\
L--D & MNLI-mm & $-18.27\pm0.32$ & $[-19.28,-17.24]$ & $-18.00\pm0.41$ & $[-19.03,-16.95]$ \\
L--D & SICK & $-2.93\pm0.26$ & $[-3.88,-1.96]$ & $-4.01\pm0.20$ & $[-5.10,-2.93]$ \\
\bottomrule
\end{tabular}
\caption{Paired differences in percentage points (three-seed mean $\pm$ sample standard deviation), with crossed paired bootstrap 95\% confidence intervals. L/T/FH/D: Linker, Transfer, FH-SAT, Direct.}
\label{tab:all-paired-results}
\end{table*}

\begin{table}[!htbp]
\centering
\footnotesize
\setlength{\tabcolsep}{2pt}
\begin{tabular}{lrrr}
\toprule
Data & 2026 & 2027 & 2028 \\
\midrule
SNLI & $<10^{-300}$ & $<10^{-300}$ & $<10^{-300}$ \\
MNLI-m & $7.67\times10^{-264}$ & $4.41\times10^{-260}$ & $4.67\times10^{-275}$ \\
MNLI-mm & $5.40\times10^{-290}$ & $3.29\times10^{-260}$ & $3.01\times10^{-293}$ \\
SICK & $8.46\times10^{-222}$ & $1.82\times10^{-28}$ & $1.34\times10^{-29}$ \\
\bottomrule
\end{tabular}
\caption{Exact McNemar $p$-values for Linker versus Transfer, Holm-corrected across four splits within each seed. Underflowed values are reported as bounds.}
\label{tab:per-seed-paired}
\end{table}

Tables~\ref{tab:main-per-seed}--\ref{tab:per-seed-paired} report per-seed scores
and paired comparisons.
Comparisons pair predictions on the same complete evaluation split within
each seed. We use 10,000 crossed paired-bootstrap replicates, resampling
seeds and evaluation examples independently. Each replicate uses the same
sampled examples for all selected seeds and both systems, then averages
their per-seed metric differences. Each neural seed is compared with the same
deterministic FH-SAT prediction vector. For each system comparison and seed,
we apply Holm correction to the exact McNemar tests across the four
evaluation splits.

\subsection{Invalid Outputs}
\label{app:invalid-efficiency}
\begin{table}[!htbp]
\centering
\small
\setlength{\tabcolsep}{3pt}
\begin{tabular}{lrrrrr}
\toprule
Data & $N$ & D & T & L & FH \\
\midrule
SNLI & 9,824 & 0 & 0 & 0 & 182 \\
MNLI-m & 9,815 & 0 & 6 & 6 & 701 \\
MNLI-mm & 9,832 & 0 & 10 & 10 & 796 \\
SICK & 4,906 & 0 & 1 & 1 & 0 \\
\bottomrule
\end{tabular}
\caption{Invalid counts on full splits. D/T/L: Direct, Transfer, Linker. Counts are identical across neural seeds; FH-SAT is deterministic. Invalid outputs count as errors.}
\label{tab:invalid-breakdown}
\end{table}

All SAT systems return Invalid for unavailable representations
(Table~\ref{tab:invalid-breakdown}). Transfer and Learned Linker also reject
an inconsistent premise--exact-match base, while FH-SAT rejects an
inconsistent premise or substituted hypothesis.

\subsection{Published Benchmark Context}
\label{app:published-comparison}
\begin{table}[!htbp]
\centering
\small
\setlength{\tabcolsep}{1.3pt}
\begin{tabular}{p{0.46\columnwidth}rr}
\toprule
Method & SNLI & SICK \\
\midrule
Differentiable Natural Logic \citep{feng-etal-2020-exploring}
  & 81.2 & -- \\
Neuro-symbolic Natural Logic + IR \citep{feng-etal-2022-neuro}
  & 87.5 & -- \\
SLR-NLI \citep{stacey-etal-2022-logical}
  & 90.33 & 85.43 \\
NeuralLog \citep{chen-etal-2021-neurallog}
  & -- & 90.3 \\
Learned Linker
  & $78.22\pm0.30$ & $84.17\pm0.24$ \\
\bottomrule
\end{tabular}
\caption{Published Accuracy (\%) under different training and evaluation
settings. IR: introspective revision. Our SICK evaluation contains 4,906 examples; NeuralLog reports
results on 4,927. The SNLI result for \citet{feng-etal-2020-exploring} is
reported by \citet{feng-etal-2022-neuro}.
Our results are three-seed means and standard deviations; a dash denotes an
unreported result.}
\label{tab:published-comparison}
\end{table}

The prior systems in Table~\ref{tab:published-comparison} do not report results under our full public MNLI
matched/mismatched protocol.

\FloatBarrier
\section{Data Scaling}
\label{app:scaling}
\begin{table}[!htbp]
\centering
\small
\setlength{\tabcolsep}{2.5pt}
\begin{tabular}{lrrr}
\toprule
Data & 30k & 60k & $\Delta$ \\
\midrule
\multicolumn{4}{l}{\emph{Accuracy}} \\
SNLI & $78.22\pm0.30$ & $79.60\pm0.78$ & $1.38\pm0.48$ \\
MNLI-m & $61.40\pm0.60$ & $62.35\pm1.18$ & $0.96\pm1.65$ \\
MNLI-mm & $62.00\pm0.39$ & $62.59\pm0.93$ & $0.59\pm1.28$ \\
\addlinespace
\multicolumn{4}{l}{\emph{Macro-F1}} \\
SNLI & $78.17\pm0.28$ & $79.54\pm0.72$ & $1.37\pm0.45$ \\
MNLI-m & $61.61\pm0.63$ & $62.52\pm1.31$ & $0.91\pm1.79$ \\
MNLI-mm & $62.17\pm0.42$ & $62.71\pm1.04$ & $0.54\pm1.40$ \\
\bottomrule
\end{tabular}
\caption{Learned Linker scaling (\%). SNLI and MNLI increase from 30,000 to 60,000 training examples. Values are three-seed means and sample standard deviations. $\Delta$: paired 60k minus 30k difference in percentage points.}
\label{tab:data-scaling}
\end{table}

For SNLI and MNLI, we enlarge the training sets from 30,000 to 60,000
examples, with 20,000 per label, using the selection procedure and seed in
Appendix~\ref{app:data-selection}. Each enlarged set contains the original
30,000 examples. We compare runs with matched training seeds, initialization,
development sets, and settings (Appendix~\ref{app:hyperparameters}).
The larger sets yield more updates per epoch.

Three-seed mean Accuracy and Macro-F1 improve on SNLI and both MNLI splits
(Table~\ref{tab:data-scaling}). Mean Accuracy gains range from 0.59 to 1.38
percentage points, although not all individual seeds improve. These
matched-seed comparisons are descriptive.

\section{Representation Complexity and Local Label Coverage}
\label{app:target-complexity}
\begin{table*}[!tp]
\centering
\small
\setlength{\tabcolsep}{4pt}
\begin{tabular}{llrrrrr}
\toprule
Data & Hyp. atoms & Examples & Acc. & M-F1 & H1 coverage & Invalid count \\
\midrule
SNLI & 1 & 3,386 & $83.19\pm0.10$ & $80.66\pm0.04$ & $99.79\pm0.00$ & 0 \\
 & 2 & 2,847 & $77.87\pm0.37$ & $77.65\pm0.37$ & $98.22\pm0.32$ & 0 \\
 & 3-4 & 2,506 & $75.01\pm0.09$ & $75.03\pm0.05$ & $96.21\pm0.60$ & 0 \\
 & 5-8 & 1,005 & $71.91\pm1.70$ & $70.13\pm1.75$ & $91.71\pm0.85$ & 0 \\
 & $\geq9$ & 80 & $60.42\pm1.91$ & $55.27\pm1.47$ & $79.17\pm1.44$ & 0 \\
\midrule
MNLI-m & 1 & 817 & $65.03\pm0.14$ & $61.57\pm0.20$ & $93.15\pm0.00$ & 0 \\
 & 2 & 1,615 & $64.44\pm0.92$ & $63.68\pm1.03$ & $90.30\pm0.39$ & 1 \\
 & 3-4 & 3,570 & $61.06\pm0.38$ & $61.36\pm0.44$ & $88.94\pm0.06$ & 1 \\
 & 5-8 & 3,272 & $60.61\pm0.88$ & $60.53\pm0.83$ & $87.20\pm0.73$ & 2 \\
 & $\geq9$ & 539 & $53.99\pm0.81$ & $52.57\pm0.86$ & $81.14\pm0.43$ & 0 \\
 & unavailable & 2 & $0.00\pm0.00$ & $0.00\pm0.00$ & $0.00\pm0.00$ & 2 \\
\midrule
MNLI-mm & 1 & 455 & $69.45\pm0.58$ & $66.62\pm0.49$ & $94.29\pm0.00$ & 0 \\
 & 2 & 1,203 & $67.55\pm0.71$ & $65.87\pm0.47$ & $94.38\pm0.34$ & 0 \\
 & 3-4 & 3,410 & $63.02\pm0.32$ & $63.04\pm0.38$ & $91.84\pm0.26$ & 0 \\
 & 5-8 & 3,939 & $60.26\pm0.70$ & $60.29\pm0.60$ & $88.20\pm0.21$ & 0 \\
 & $\geq9$ & 815 & $54.56\pm0.31$ & $54.31\pm0.25$ & $82.17\pm0.68$ & 0 \\
 & unavailable & 10 & $0.00\pm0.00$ & $0.00\pm0.00$ & $0.00\pm0.00$ & 10 \\
\midrule
SICK & 1 & 1,411 & $86.72\pm0.18$ & $86.88\pm0.19$ & $97.45\pm0.00$ & 0 \\
 & 2 & 1,185 & $83.97\pm0.66$ & $82.49\pm0.72$ & $96.96\pm0.22$ & 0 \\
 & 3-4 & 1,573 & $81.06\pm0.50$ & $76.60\pm0.58$ & $95.36\pm0.06$ & 0 \\
 & 5-8 & 706 & $86.59\pm0.87$ & $80.40\pm0.73$ & $95.18\pm0.65$ & 0 \\
 & $\geq9$ & 31 & $78.49\pm1.86$ & $62.51\pm3.29$ & $93.55\pm0.00$ & 1 \\
\bottomrule
\end{tabular}
\caption{Learned Linker performance and complete Hamming-1 gold-label coverage by hypothesis atom count (\%). Accuracy, Macro-F1 and coverage are three-seed means and sample standard deviations; Invalid counts are identical across seeds. Invalid outputs count as errors; unavailable representations form a separate row.}
\label{tab:target-atom-complexity}
\end{table*}

We group the main experiment's full-split single-MAP Linker predictions
by hypothesis atom count (Table~\ref{tab:target-atom-complexity}). Invalid
outputs count as errors; unavailable representations form a separate bucket
and remain in the overall denominator.

For each selected checkpoint, evaluation-set Hamming-1 coverage is the
fraction of examples whose gold label is produced by the MAP world or a legal
one-target alternative. This measures gold-label availability, not link correctness.

Mean Accuracy decreases across hypothesis-size buckets on SNLI and both
MNLI splits but is nonmonotonic on SICK. Bucket sizes and label distributions
differ, so these associations do not isolate a causal effect of representation
size.
The largest-atom SNLI and SICK buckets contain only 80 and 31 examples,
respectively.

Overall evaluation-set coverage is SNLI: $97.43\pm0.31\%$; MNLI-m: $88.49\pm0.34\%$; MNLI-mm: $89.91\pm0.27\%$; SICK: $96.31\pm0.09\%$.

\paragraph{Training-set coverage.}
\begin{table}[!htb]
\centering
\small
\setlength{\tabcolsep}{4pt}
\begin{tabular}{lrr}
\toprule
Data & Epoch 1 & Selected epoch \\
\midrule
SNLI & $74.78\pm21.56$ & $97.24\pm0.74$ \\
MNLI & $58.11\pm28.67$ & $89.37\pm0.18$ \\
SICK & $62.75\pm30.44$ & $97.14\pm0.27$ \\
\bottomrule
\end{tabular}
\caption{Training-set Hamming-1 gold-label coverage (\%) before updates at epoch 1 and each run's development-selected epoch. Values are three-seed means and sample standard deviations.}
\label{tab:training-coverage}
\end{table}

For the main training sets, Table~\ref{tab:training-coverage} reports the
fraction of examples whose frozen Hamming-1 pool contains a world with the
gold label. Coverage does not imply a nonzero loss or gradient.

\FloatBarrier
\section{Structural Workload and Computational Cost}
\label{app:efficiency}
\begin{table*}[!tp]
\centering
\small
\setlength{\tabcolsep}{4pt}
\begin{tabular}{lrrrr}
\toprule
Quantity & SNLI & MNLI-m & MNLI-mm & SICK \\
\midrule
Premise atoms $|\mathcal A_P|$ & 5.7 / 5.0 / 12.0 & 9.2 / 8.0 / 21.0 & 9.6 / 8.0 / 22.0 & 2.8 / 2.0 / 6.0 \\
Hypothesis atoms $|\mathcal A_H|$ & 2.5 / 2.0 / 6.0 & 4.3 / 4.0 / 9.0 & 4.9 / 4.0 / 10.0 & 2.7 / 2.0 / 6.0 \\
$|\mathcal A_P||\mathcal A_H|$ & 15.0 / 10.0 / 45.0 & 45.0 / 30.0 / 143.0 & 52.9 / 36.0 / 154.0 & 9.8 / 6.0 / 32.0 \\
Candidate pairs & 12.4 / 8.0 / 36.0 & 35.0 / 21.0 / 112.0 & 40.6 / 27.0 / 120.0 & 6.5 / 4.0 / 24.0 \\
$|\Wlocal|$ & 25.9 / 17.0 / 73.0 & 71.1 / 43.0 / 225.0 & 82.3 / 55.0 / 241.0 & 14.0 / 9.0 / 49.0 \\
$\log_{10}|\mathcal W|$ & 2.2 / 1.7 / 5.7 & 4.2 / 3.5 / 10.2 & 4.8 / 4.2 / 11.1 & 1.6 / 1.0 / 4.8 \\
\bottomrule
\end{tabular}
\caption{Structural workload for available representations: mean / median / 95th percentile (nearest-rank). Values are independent of the training seed.}
\label{tab:representation-workload}
\end{table*}

Table~\ref{tab:representation-workload} reports counts of all atom pairs
$|\mathcal A_P||\mathcal A_H|$, candidate pairs $|\mathcal A_P||\mathcal U|$,
local worlds $|\Wlocal|$, and the base-10 logarithm of the complete-world
count, $\log_{10}|\mathcal W|$.
Exact-matched targets are excluded from candidate pairs.

\begin{table}[!htbp]
\centering
\small
\setlength{\tabcolsep}{4pt}
\begin{tabular}{llrr}
\toprule
Data & System & \shortstack{Completed\\epochs} & Minutes \\
\midrule
SNLI & Direct Sentence & 9/8/9 & $12.52\pm0.81$ \\
 & Learned Linker & 10/10/8 & $55.62\pm6.16$ \\
MNLI & Direct Sentence & 6/7/6 & $9.30\pm0.83$ \\
 & Learned Linker & 10/10/10 & $160.72\pm4.78$ \\
SICK & Direct Sentence & 16/16/20 & $3.07\pm0.40$ \\
 & Learned Linker & 20/20/20 & $11.58\pm0.34$ \\
\bottomrule
\end{tabular}
\caption{Recorded training times for the main experiments; resource contention was not controlled. Completed training epochs follow seed order 2026/2027/2028; minutes are mean $\pm$ sample standard deviation of recorded run totals.}
\label{tab:training-efficiency}
\end{table}

\begin{table}[!htbp]
\centering
\footnotesize
\setlength{\tabcolsep}{2.5pt}
\begin{tabular}{llrr}
\toprule
Split & System & Seconds & Examples/s \\
\midrule
SNLI & Direct & $9.04\pm0.07$ & $1087.26\pm8.08$ \\
 & Transfer & $109.85\pm0.12$ & $89.43\pm0.10$ \\
 & Linker & $23.15\pm0.83$ & $424.65\pm15.04$ \\
MNLI-m & Direct & $9.16\pm0.01$ & $1071.24\pm1.45$ \\
 & Transfer & $306.04\pm0.52$ & $32.07\pm0.05$ \\
 & Linker & $61.65\pm1.25$ & $159.25\pm3.25$ \\
MNLI-mm & Direct & $9.19\pm0.06$ & $1070.37\pm6.54$ \\
 & Transfer & $359.51\pm0.52$ & $27.35\pm0.04$ \\
 & Linker & $71.13\pm1.33$ & $138.26\pm2.56$ \\
SICK & Direct & $4.68\pm0.03$ & $1048.65\pm6.37$ \\
 & Transfer & $29.01\pm0.13$ & $169.10\pm0.76$ \\
 & Linker & $7.06\pm0.25$ & $695.01\pm24.95$ \\
\bottomrule
\end{tabular}
\caption{Seed-2026 inference times on one RTX 4090 (three-run mean $\pm$ sample SD). Timing boundaries and padding are specified in Appendix~\ref{app:efficiency}.}
\label{tab:efficiency}
\end{table}

Training times in Table~\ref{tab:training-efficiency} are recorded totals for
the main experiments, including development evaluation and early stopping.
Seed-2026 totals include only the selected
search runs. Epoch counts and stopping times vary, so these totals do not
measure training speed under matched conditions.

\begin{samepage}
Inference times in Table~\ref{tab:efficiency} are measured on one RTX 4090 with fixed seed-2026 checkpoints and
four CPU threads. Transfer retains Direct Sentence's fixed 256-token padding;
Learned Linker uses 64 tokens.
Direct Sentence batches 64 sentence pairs.
The atom systems batch at most 16 examples and 128 candidate atom pairs;
each example exceeding the pair limit is processed alone.
Each system runs one untimed warmup
and three timed full-split repetitions. Systems run sequentially, with CUDA
synchronized before and after each timed run.

\end{samepage}

\emph{Symbolic preprocessing.} We construct CNF clauses, exact matches, and
legal link candidates from cached representations. This preprocessing is
timed separately and excluded from Table~\ref{tab:efficiency}:
SNLI 15.04s, MNLI-m 31.45s, MNLI-mm 34.52s, and SICK 5.25s.

\emph{Measured inference.} Timing includes checkpoint and model loading, assembling and indexing
text-pair inputs, tokenization, neural scoring, and SAT-based inference.
The atom systems construct these text-pair inputs from the precomputed candidates.
Neural predictions and SAT results are not reused across runs;
operating-system file caches remain warm.
Timing excludes AMR parsing, representation compilation, one-time loading of
evaluation sentence pairs and cached propositional representations, and
post-run verification.

\FloatBarrier
\section{Implementation Details}
\label{app:link-construction}

\paragraph{Parsing and atom verbalizations.}
We produce AMRs with the transition-AMR parser's
\texttt{AMR3-joint-ontowiki-seed42} checkpoint \citep{lee-etal-2022-maximum}, using
\texttt{transition-neural-parser} version 0.5.4.
Atom-pair NLI uses the translator's verbalizations unchanged.

\paragraph{Deterministic ordering.}
Witness ties select the first world in a fixed order: the epoch-start MAP world,
then one-target changes in atom-identifier order and the action order in
Equation~\ref{eq:legal-actions}.

\paragraph{SAT reasoner queries.}
SymPy 1.12 converts $\Phi_P$ and $\Phi_H$ to equivalent conjunctive normal form
(CNF) without auxiliary Boolean atoms \citep{meurer-etal-2017-sympy}.
Appending the link clauses gives a CNF of $\Phi_w$. Glucose3 through
\texttt{python-sat} 0.1.8.dev9 executes the queries in
Definition~\ref{def:world-label} \citep{ignatiev-etal-2018-pysat}.
For each clause $\gamma$ in the CNF of $\Phi_H$, we query the CNF of $\Phi_w$
under assumptions
$\{\neg\ell:\ell\in\gamma\}$, which make every literal in $\gamma$ false.
The world entails $\Phi_H$ exactly when every such query is unsatisfiable.

\paragraph{Licenses.}
Our original code is released under the MIT License. SNLI-derived artifacts
follow CC BY-SA 4.0, and SICK-derived artifacts follow CC BY-NC-SA 3.0.
MultiNLI materials retain their source-specific terms. Bundled model resources
retain their upstream MIT notices. The repository lists licenses,
attribution, and preprocessing.

\end{document}